\documentclass[runningheads]{llncs}

\usepackage[mobile]{eccv}

\usepackage{eccvabbrv}

\usepackage{graphicx}
\usepackage{booktabs}

\usepackage[accsupp]{axessibility}  

\usepackage{hyperref}

\usepackage{orcidlink}

\usepackage{wrapfig}
\definecolor{cvprblue}{rgb}{0.12,0.49,0.85}
\definecolor{visualblue}{RGB}{3,5,216}
\definecolor{visualgreen}{RGB}{0,250,2}
\definecolor{visualred}{RGB}{220,3,12}
\definecolor{sm}{RGB}{34, 98, 186}
\definecolor{resultcolor}{RGB}{236,242,250}
\usepackage{array}
\usepackage[table]{xcolor}
\usepackage{multirow}
\usepackage[most]{tcolorbox}
\usepackage{bbold}
\usepackage{graphicx} 
\usepackage{amsmath}
\usepackage{algorithm}
\usepackage{algorithmic}
\usepackage{titletoc}

\begin{document}

\title{Agentic Collaborative Cognition for Zero-Shot 3D Understanding} 

\titlerunning{Agentic Collaborative Cognition for Zero-Shot 3D Understanding}

\author{
Wenxin Wang$^{1}$\thanks{Equal contribution. \quad $\dagger$ Project lead. \quad $\ddagger$ Corresponding author.},
Bo Zhang$^{1\star}$,
Feng Chen$^{2\dagger}$,
Zixuan Wang$^{1}$,
Wen Li$^{3}$,
Changsheng Li$^{4}$,
Yinjie Lei$^{1\ddagger}$,
}

\authorrunning{W. Wang et al.}

\institute{\small
$^{1}$Sichuan University, $^{2}$Adelaide University, 
\\
$^{3}$University of Electronic Science and Technology of China,
\\
$^{4}$Beijing Institute of Technology
\\
{
\email{2024222050082@stu.scu.edu.cn}, \email{yinjie@scu.edu.cn}
}
 }

\maketitle

\begin{abstract}
  Recent advancements have explored agentic zero-shot 3D understanding by reformulating it as video keyframe understanding with Multimodal Large Language Models (MLLMs). However, existing methods face an intrinsic bottleneck due to the finite observation perspectives inherent in videos and the implicit perception of 3D scenes. In this paper, we propose a collaborative multi-agent framework that assigns a Planning Agent to handle high-level viewpoint planning and supplement novel perspectives, and a Perception Agent to explicitly summarize the 3D scene into a structured holistic cognitive map. Specifically, Planning Agent first analyzes this cognitive map to determine query-relevant viewpoints and supplements missing critical perspectives to ensure comprehensive observation. Subsequently, Perception Agent documents object-level attributes from these views by assigning consistent instance identifiers across viewpoints, thereby integrating fragmented observations into the holistic cognitive map. In parallel, it provides feedback to filter out mismatched candidate objects and guide subsequent viewpoint planning. Through this closed-loop iterative process, two agents collaboratively figure out candidates until Perception Agent determines that sufficient information has been captured to complete the task. Extensive experiments demonstrate that our method achieves state-of-the-art performance on 6 benchmarks, with improvements of 11.1\% Acc@0.5 on ScanRefer, 14.6 BLEU-1 on 3D-assisted dialog, and 2.1 EM on SQA3D. \noindent\textbf{Project Page}: \url{https://zhangbo135.github.io/agentic-collaborative-cognition/}
  \keywords{Zero-Shot 3D Understanding \and Collaborative Cognition \and Self-evolving Agent}
\end{abstract}

\section{Introduction}
\label{sec:intro}

3D scene understanding~\cite{zhang2024towards, jin2023context, wu2025spatial} focuses on interpreting spatial layouts, semantic attributes, and inter-object relationships in complex environments, enabling critical applications from robotics~\cite{wang2019reinforced, xia2018gibson} to augmented reality~\cite{liu2023segment, arena2022overview}. Previous studies mainly rely on fully supervised paradigms to achieve 3D visual grounding~\cite{chen2020scanrefer, zheng2025video}, 3D dense captioning~\cite{chen2021scan2cap, huang2024chat}, 3D question answering~\cite{azuma2022scanqa, ma2022sqa3d}, \textit{etc}. Despite their effectiveness, the heavy reliance on large-scale 3D-text paired datasets limits scalability and adaptability to diverse real-world environments~\cite{yang2023zero, yuan2024visual}.

\begin{figure*}[t]
\centering
\includegraphics[width=\linewidth]{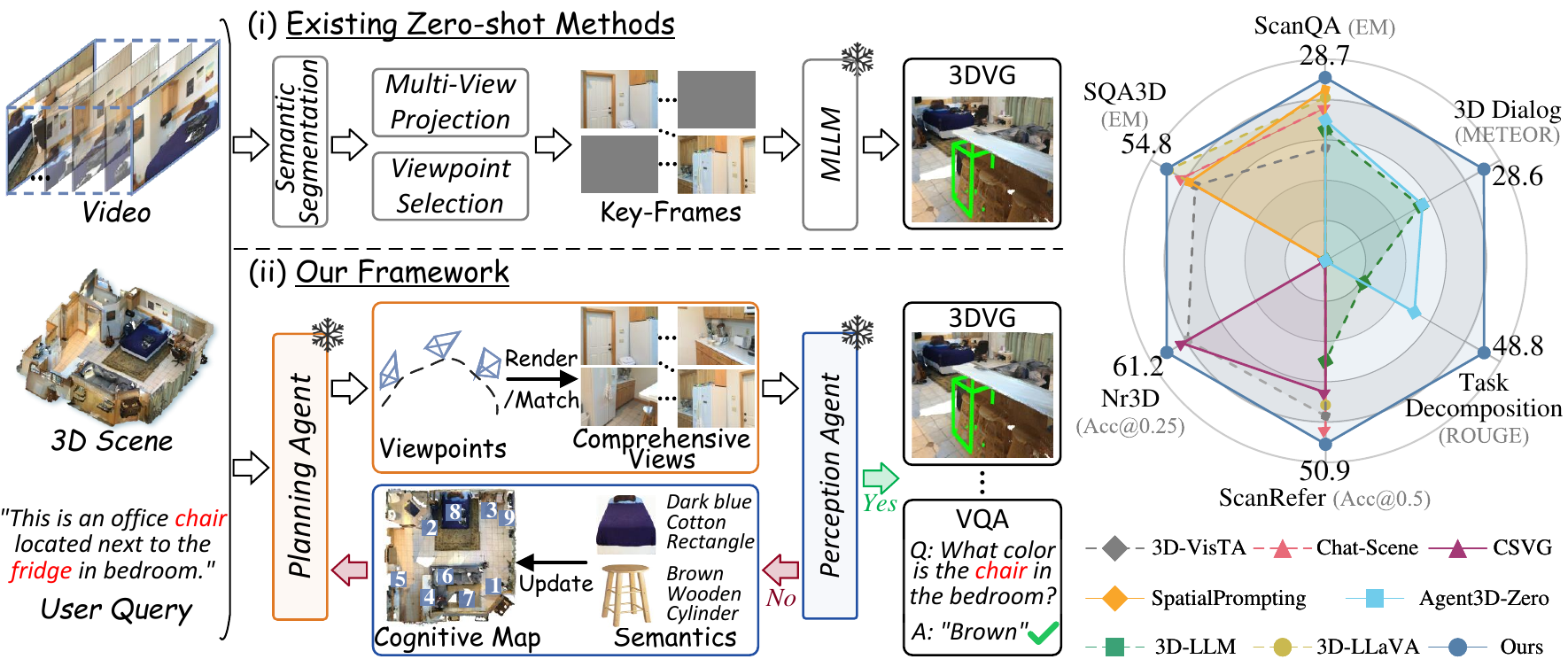} 
\caption{\textbf{Comparison of existing methods with our method.} Left: Existing zero-shot methods~\cite{jin2025spazer, lin2025seqvlm} typically rely on video keyframe selection for 3D visual grounding. Our method actively plans viewpoints to obtain comprehensive views and explicitly summarizes the 3D scene into a structured holistic cognitive map from these views for diverse 3D tasks. Right: Our method achieves substantial improvements over existing zero-shot and most fully supervised methods across 6 benchmarks.} 
\label{fig:abstract}
\end{figure*}

Zero-shot 3D understanding~\cite{zhang2024agent3d, yuan2024solving, taguchi2025spatialprompting} has emerged to reduce the requirement for manually annotated datasets~\cite{dai2017scannet, ramakrishnan2021habitat}. As shown in the top-left of \cref{fig:abstract}, previous methods~\cite{jin2025spazer, lin2025seqvlm} typically leverage the general world knowledge of Multimodal Large Language Models (MLLMs)~\cite{qwen2vl, hurst2024gpt} to interpret keyframes extracted from scene videos (an inherent component of scanned 3D datasets) through object-centric keyframe selection, which is enabled by key proposal selection and multi-view projection. However, relying solely on video keyframe selection inevitably has a performance bottleneck because: (1) \textit{Finite observation perspectives.} During 3D scene scanning, videos are typically captured from restricted viewpoints close to the scene surfaces to ensure reconstruction quality. However, they provide only finite frames with fixed viewpoints, which often lack query-relevant critical perspectives; (2) \textit{Implicit spatial perception.} Existing methods~\cite{jin2025spazer, taguchi2025spatialprompting} typically leverage MLLMs to implicitly infer the spatial layout from keyframes. Although capable of capturing inter-object relationships under similar viewpoints~\cite{zheng2025video, qi2025gpt4scene, wang2025ross3d}, they struggle to maintain unified spatial cognition when facing significant viewpoint variations~\cite{yang2025thinking, xie2025dsm}.

In this paper, we propose a collaborative multi-agent framework for zero-shot 3D scene understanding. As illustrated in the bottom-left of \cref{fig:abstract}, our method consists of two specialized agents: a Planning Agent that performs flexible planning of query-relevant viewpoints and supplements missing critical perspectives, and a Perception Agent that explicitly summarizes the 3D environment into a structured holistic cognitive map. Specifically, Planning Agent first analyzes this cognitive map to retrieve critical candidate objects and determine query-relevant viewpoints to further figure out these objects. For missing critical perspectives, it supplements rendered views to ensure comprehensive observation for Perception Agent. Subsequently, Perception Agent assigns unique instance identifiers to all objects in each view to ensure cross-view consistency, thereby accurately documenting object appearance and physical attributes, and integrating originally fragmented observations into the cognitive map. Additionally, it provides feedback that filters out mismatched objects to narrow the scope of candidates and guides the next round of viewpoint planning. This iterative collaboration allows the two agents to progressively refine 3D scene cognition until Perception Agent determines that sufficient information has been captured.

Empirically, as illustrated in the right of \cref{fig:abstract}, our method surpasses previous state-of-the-art zero-shot and most fully supervised methods across 6 benchmarks, achieving substantial improvements of 11.1\% Acc@0.5 on ScanRefer~\cite{chen2020scanrefer}, 14.6 BLEU-1 on 3D-assisted dialog~\cite{hong20233d}, and 2.1 EM on SQA3D~\cite{ma2022sqa3d}. In brief, our main contributions can be summarized as follows:

\begin{itemize}

\item We propose a collaborative multi-agent framework that reformulates zero-shot 3D understanding as an interactive planning-perception process.

\item We design a Planning Agent to plan query-relevant viewpoints for comprehensive observation, and a Perception Agent to update the holistic cognitive map, enabling the MLLM to directly observe and understand 3D scenes.

\item Extensive experiments on 6 benchmarks validate the state-of-the-art performance and generality of our method across diverse 3D understanding tasks.

\end{itemize}

\section{Related work}
\label{sec:Related_work}

\noindent\textbf{Zero-shot 3D Scene Understanding.} 
Zero-shot 3D scene understanding has been explored as a promising alternative to fully supervised paradigms. Early works \cite{yang2024llm, gu2024conceptgraphs, yuan2024visual} typically reformat the 3D scene into textual descriptions for LLMs. For example, LLM-Grounder~\cite{yang2024llm} uses the LLM to decompose queries into semantic constituents and employs a grounding tool to identify the target. CSVG~\cite{yuan2024solving} constructs structured scene graphs to model inter-object spatial relationships. However, these methods rely solely on text and struggle to capture rich visual details and fine-grained cues. Recent studies~\cite{xu2024vlm, lin2025seqvlm, taguchi2025spatialprompting} leverage MLLMs to understand the 3D scene from 2D videos. SeqVLM~\cite{lin2025seqvlm} develops a proposal-guided projection strategy that performs object-centric keyframe selection to localize the target object. SPAZER~\cite{jin2025spazer} proposes a coarse-to-fine progressive reasoning mechanism to enable 3D-2D joint decision-making. Despite their effectiveness, they remain limited by the finite observation perspectives and fail to explicitly model the entire 3D scene.

\noindent\textbf{3D Scene Documenting.}
Effective 3D scene documenting is a key approach to enhance understanding capabilities, evolving from geometric maps to rich semantic structures~\cite{rosinol20203d, mascaro2024scene}. SceneGraphFusion~\cite{wu2021scenegraphfusion} proposes a real-time method that incrementally constructs a globally consistent representation to document the 3D scene. More recently, Clio~\cite{maggio2024clio} leverages LLMs to construct language-based open-vocabulary 3D scene graphs. MTU3D~\cite{zhu2025move} uses FastSAM~\cite{zhao2023fast} and DINO~\cite{oquab2023dinov2} to implicitly construct a dynamic spatial memory bank. Embodied VideoAgent~\cite{fan2025embodied} analyzes egocentric video and embodied sensory inputs to build a persistent scene memory. However, these representations often focus on categorical labels and coarse scene-level state, lack explicit multi-dimensional attributes, and typically require several tools to query the memory, whereas our cognitive map provides a flexible and fine-grained representation and further serves as a shared state for inter-agent communication.

\noindent\textbf{MLLM-Based Multi-Agent Collaboration.}
With the rapid advancement of MLLMs~\cite{bai2023qwen, hurst2024gpt}, recent research has increasingly explored a wide range of applications that focus on MLLM-based multi-agent collaboration for multimodal reasoning. This allows multiple agents to work together, which enables the collective intelligence to efficiently solve complex and multi-step tasks via explicit role specialization and communication. For example, leveraging MLLMs for role-playing can accomplish multiple challenging tasks, such as embodied navigation~\cite{shen2025enhancing}, social simulation~\cite{liu2024lmagent, vera2025multimodal}, video understanding~\cite{zhou2025reagent}, and video generation~\cite{yuan2024mora}. This paradigm is particularly relevant to 3D scene understanding, which naturally involves two heterogeneous sub-tasks: viewpoint acquisition and perceptual verification. Relying on a single agent inevitably leads to functional entanglement, causing insufficient verification and potential mislocalization of the target object. Therefore, we adopt a collaborative multi-agent framework that explicitly decouples viewpoint planning from perceptual verification.






\section{Methodology}
\label{sec:method}

Given a query, the goal of 3D scene understanding is to either localize the referring target object or answer questions about object attributes and spatial layout. In this work, we propose a general collaborative multi-agent framework that progressively refines the cognition of the 3D scene through iterative interactions between planning and perception for diverse 3D understanding tasks. Here are two key benefits of our method: (1) Rather than relying on static object-centric keyframe selection from the video~\cite{lin2025seqvlm, jin2025spazer}, we actively identify critical objects, capture their relationships with surrounding objects, and allow targeted observations of uncertain objects, ensuring that query-relevant information is fully captured; (2) The holistic cognitive map serves as an explicit and structured documentation of the 3D scene, shared among agents. By integrating spatial and semantic information, it facilitates consistent understanding and effective collaboration between agents.

\subsection{Overview}
\label{sec:3.1}

\begin{figure*}[t]
\centering
\includegraphics[width=\linewidth]{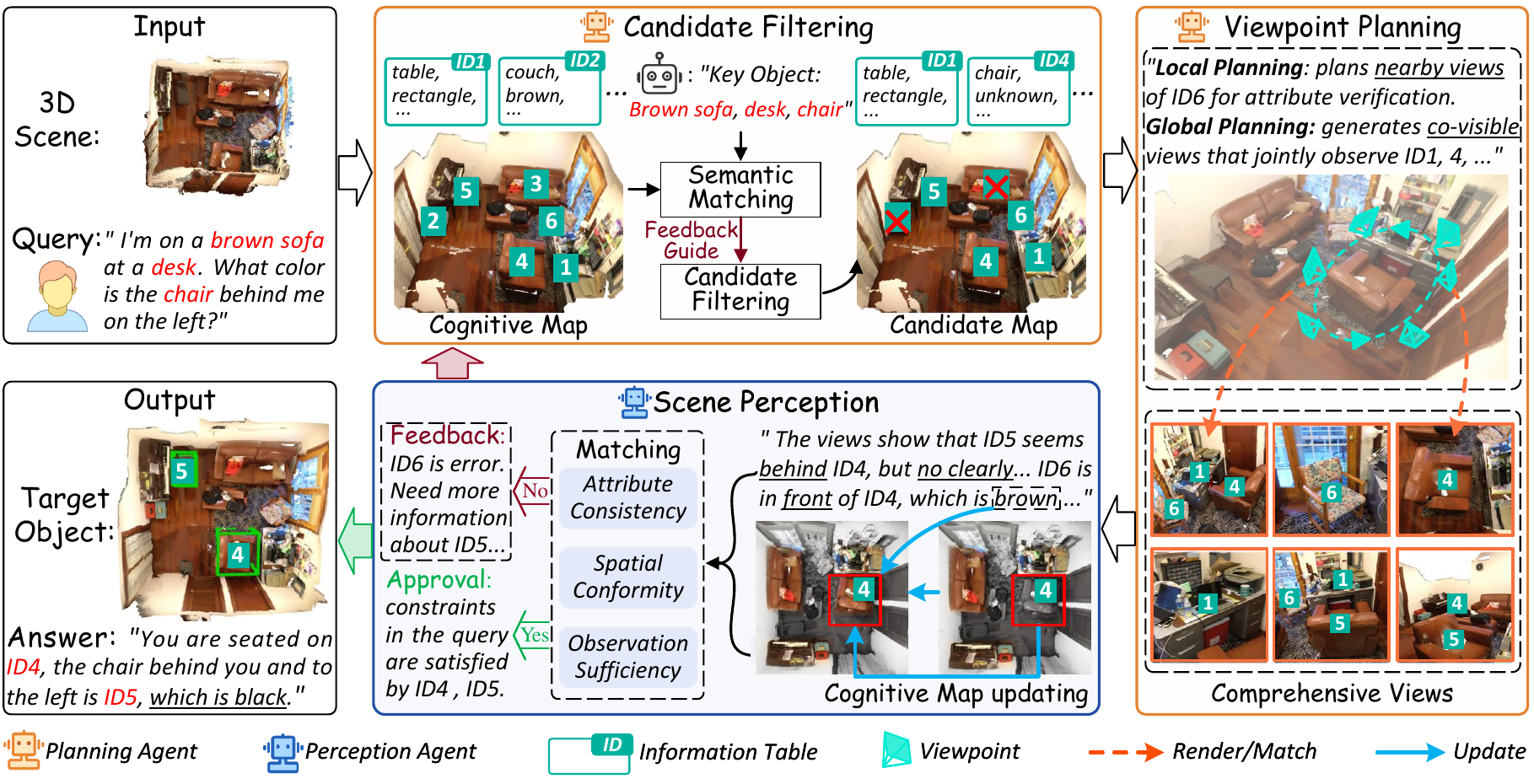} 
\caption{\textbf{Overview of our method.} Given a 3D scene and query, our method achieves zero-shot 3D understanding through collaboration of planning and perception agents. We first construct a holistic cognitive map that contains an information table and a BEV annotated with instance identifiers. Planning Agent conducts candidate filtering to retrieve query-relevant objects, plans viewpoints and maps them to obtain comprehensive views. Perception Agent then perceives the scene from these views, and updates this map with documented object attributes. It matches current observations with query, and provides feedback to refine candidates or generates answer.
}
\label{fig:pipeline}
\end{figure*}

The overall framework is illustrated in \cref{fig:pipeline}. Given a textual query ${Q}$, our method incorporates a Planning Agent and a Perception Agent to collaboratively capture query-relevant scene information. Based on this, we first initialize a holistic cognitive map $\mathcal{M} = \{\mathcal{T}, {I}_{B}\}$ that explicitly summarizes the scene. $\mathcal{T} = \{\sigma_i\}_{i=1}^{N}$ is an object information table that stores object-level information $\sigma_i = \{b_i, l_i, a_i\}$, where $b_i$ and $l_i$ represent the 3D bounding box and category label extracted by a 3D detection framework~\cite{schult2023mask3d}, and $a_i$ contains attributes of each object. At initialization, $a_i$ is empty and will be progressively updated by Perception Agent. $\mathcal{T}$ can be converted into a text format compatible with the MLLM input, providing an accurate spatial and semantic description of the 3D scene. $I_B$ is a bird’s-eye view (BEV), a 2D projection of the scene, annotated with the instance identifier $i$ of each object.

We first parse the textual query ${Q}$ to identify the key object set $\mathcal{O}=\{o_j\}_{j=1}^{J}$ of all objects involved in ${Q}$, where $J$ denotes the number of key objects. During the ${t}$-th interaction round, Planning Agent first filters out irrelevant and erroneous candidate objects from $\mathcal{M}$ to obtain the candidate map $\mathcal{C}^t$, and then plans a set of viewpoints $\mathcal{V}^{t}$ to further figure out candidates. These viewpoints $\mathcal{V}^{t}$ are then matched with real images or supplemented with rendered images to obtain comprehensive observations $\mathcal{I}^{t}$. For each image in $\mathcal{I}^{t}$, we annotate all candidate objects with their unique identifiers, serving as visual prompts to ensure identity consistency across different viewpoints. Based on the annotated images, Perception Agent performs fine-grained perception, including object recognition, attribute extraction, and spatial relationships reasoning. The extracted attributes of each object are integrated into $\mathcal{T}$ of $\mathcal{M}$ to refine the description of the 3D scene. Subsequently, Perception Agent provides feedback $\mathcal{F}^{t+1}$, which indicates erroneous candidate objects and guides the next round of viewpoint planning of Planning Agent. This iterative process continues until Perception Agent determines that attribute consistency, spatial conformity, and observation sufficiency have been achieved to generate the final answer. The complete algorithm is provided in \textcolor{cvprblue}{Appendix A.1}.

\subsection{Planning Agent}
\label{sec:3.2}
To overcome the limitations of previous methods~\cite{jin2025spazer, lin2025seqvlm} that rely on finite perspectives in the video, we introduce a Planning Agent that supplements missing critical perspectives. As shown in \cref{fig:pipeline}, Planning Agent first leverages a candidate filtering strategy to retrieve candidate objects. Subsequently, it analyzes the candidate map to actively figure out these objects through flexible viewpoint planning, which enables broader and more diverse viewpoint coverage. These planned viewpoints are then mapped to obtain comprehensive views.

\noindent\textbf{Candidate Filtering.}
We propose a retrieval-augmented candidate filtering strategy to progressively filter out irrelevant and erroneous candidate objects. For each key object ${o_j} \in \mathcal{O}$ in the query, we represent its semantics as ${o_j} = (l^q_j, a^q_j)$, consisting of the target category and attributes. We define a matching function $\operatorname{Match}(\sigma_i, o_j)$ to determine whether a scene object $\sigma_i \in \mathcal{T}$ matches $o_j$: 
\begin{equation}
\operatorname{Match}(\sigma_i, o_j)=
\begin{cases}
\mathbb{1}\!\big[l_i = l_j^{q}\big], 
& a_i=\varnothing \;\lor\; a_j^{q}=\varnothing,\\[2pt]
\mathbb{1}\!\big[l_i = l_j^{q}\big]\cdot \mathbb{1}\!\big[a_i = a_j^{q}\big], 
& \text{otherwise},
\end{cases}
\end{equation}
where $\mathbb{1}\!\big[ \cdot \big]$ denotes the indicator function and $\varnothing$ denotes that the attribute is undefined or missing. By applying this matching rule, we effectively filter out mismatched scene objects from the holistic cognitive map, thereby obtaining the initial candidate map:
\begin{equation}
\mathcal{C}^{0} = \left\{ \sigma_i \in \mathcal{T} \mid \exists\, o_j \in \mathcal{O},\ \operatorname{Match}(\sigma_i, o_j) = 1 \right\}.
\end{equation}

Unlike previous methods~\cite{jin2025spazer, lin2025seqvlm} that rely solely on object categories for retrieval, we leverage the fine-grained semantics provided by the holistic cognitive map to additionally incorporate object attributes into this matching process.

During the interaction, Planning Agent leverages feedback from Perception Agent, which identifies erroneous candidates that conflict with the query in terms of attributes or spatial relationships (as elaborated in the following \cref{sec:3.3}), to progressively refine $\mathcal{C}^{0}$ and focus on critical objects for subsequent viewpoint planning. After the ${t}$-th interaction round, $\mathcal{C}^{0}$ is refined to $\mathcal{C}^{t}$.

\noindent\textbf{Viewpoint Planning.} 
To further figure out candidate objects, Planning Agent plans a set of viewpoints based on the candidate map. Specifically, by jointly considering the query and the fine-grained guidance provided by the perceptual feedback, Planning Agent plans viewpoints close to the target objects to observe and verify their attributes, and viewpoints that simultaneously observe multiple objects to directly capture spatial relationships. These viewpoints constitute a set $\mathcal{V}^{t} = \{ {V^t_k} \}_{k=1}^{K}$, where $K$ denotes the number of planned viewpoints. Each viewpoint ${V^t_k}$ is defined by its position and orientation on the 2D BEV plane, and is subsequently converted into a 3D camera pose represented by a rotation matrix ${R}_{k}^{t}$ and a translation vector ${T}_{k}^{t}$.

Then each planned viewpoint is mapped to a real camera image or a rendered image: real images capture rich visual details but are limited by fixed camera viewpoints, whereas rendered images allow flexible observations at the cost of geometric and textural inaccuracies. Specifically, we design a viewpoint matching strategy that determines whether each planned viewpoint can be reliably matched with any real camera image. If a match is found, the corresponding real camera image is used; otherwise, a rendered image is synthesized to supplement the missing perspective and provide additional visual evidence.

Inspired by~\cite{taguchi2025spatialprompting}, we compute the pose distance $D_{k,m}^{t}$ for each planned viewpoint ${V^t_k}$ with ${T}_{k}^{t}$ and ${R}_{k}^{t}$ to each real image $I_{m}$ in the set $\mathcal{I} = \{ I_{m} \}_{m=1}^{M}$, where $M$ denotes the number of real images:

\begin{equation}
    D_{k,m}^{t} = \frac{1}{\theta } \cdot \arccos\left( \frac{\operatorname{Tr}(({R}_{k}^{t})^{\top} {R}_m) - 1}{2} \right) + \frac{1}{d} \cdot \left\| {T}_k^{t} - {T}_m \right\|_2,
    \label{eq:distance}
\end{equation}  
where ${R}_m$ and ${T}_m$ denote the rotation matrix and translation vector of the $m$-th real camera image, $\operatorname{Tr}(\cdot)$ denotes the matrix trace operator, $\theta$ and $d$ are normalization factors corresponding to the maximum rotation and translation difference. The corresponding image $I_k^{t}$ is given by:
\begin{equation}
    {I_k^{t}} =
    \begin{cases}
        I_{\hat{m}}, & \text{if } D_{k,\hat{m}}^{t} < \tau_D ,\\
        \operatorname{Render}(\mathcal{P}, {R}_k^{t}, {T}_k^{t}), & \text{otherwise},
    \end{cases}
    \quad
    \label{eq:img_match}
\end{equation}
where $\hat{m} = \mathop{\arg\min}\limits_{m=1,\ldots,M}\ D_{k,m}^{t}$ and ${\tau_D}$ is the viewpoint matching threshold. Finally, these matched and rendered 2D images are aggregated into a comprehensive view set $\mathcal{I}^{t} = \{I_k^{t} \}_{k=1}^{K}$ for Perception Agent.

\subsection{Perception Agent}
\label{sec:3.3}
The initialized holistic cognitive map contains only object layout information and lacks fine-grained attributes, such as color, shape, or material, which are crucial for precise scene understanding. To this end, as shown in \cref{fig:pipeline}, we introduce a Perception Agent that continuously documents object attributes, figures out candidate objects, and generates the final answer.

\noindent\textbf{Scene Perception.} 
We annotate each candidate object with its unique instance identifier for each image in $\mathcal{I}^{t}$, thereby ensuring cross-view consistent understanding. Subsequently, Perception Agent interprets $\mathcal{I}^{t}$ and ${\mathcal{C}}^{t}$ to figure out the candidate objects and extract the appearance attributes and physical attributes of each object. Considering the domain gap between real and rendered images, Perception Agent is prompted to prioritize extracting fine-grained semantic attributes from real images, while using rendered images primarily to infer inter-object relationships and spatial layout. These extracted attributes are dynamically integrated into $\mathcal{T}$, progressively updating and enriching the holistic cognitive map. Importantly, each attribute is assigned a confidence score, and attributes with higher confidence will overwrite previous ones, thereby identifying and correcting errors inferred from previous perceptions.

Perception Agent matches the current observations with respect to the query using three criteria: (1) Attribute conformity. Check whether the attributes of the candidates match the query; (2) Spatial consistency. Verify whether the spatial relations of the candidates satisfy the query constraints; (3) Observation sufficiency. Determine whether the current observations provide sufficient evidence to resolve the query. If any condition fails, Perception Agent provides feedback $\mathcal{F}^{t+1}$, which either identifies mismatched candidates in the case of attribute or spatial conflicts, or provides guidance for subsequent planning when observations are insufficient. Leveraging this guidance, Planning Agent actively seeks new viewpoints to resolve spatial ambiguities or capture fine-grained attribute evidence, thereby focusing on critical objects and avoiding exhaustive observations of the entire scene in an indiscriminate manner. 

Through this closed-loop iterative process, the two agents collaboratively figure out the complex 3D environment until Perception Agent confirms that the above conditions are satisfied to generate the final answer.

\section{Experiments}
\label{sec:exp}

\subsection{Experimental Settings}
\label{sec:4.1}

\noindent \textbf{Benchmarks.}
To evaluate the effectiveness of our method, we conduct extensive experiments on 6 benchmarks covering diverse 3D scene understanding tasks: 3D visual grounding on ScanRefer~\cite{chen2020scanrefer} and Nr3D~\cite{achlioptas2020referit3d}, situation estimation on SQA3D~\cite{ma2022sqa3d}, 3D question answering on ScanQA~\cite{azuma2022scanqa} and SQA3D~\cite{ma2022sqa3d}, 3D-assisted dialog and task decomposition on 3D-LLM held-in dataset~\cite{hong20233d}.

\noindent \textbf{Evaluation Metrics.}
For 3D visual grounding, we evaluate grounding accuracy using Acc@0.25 and Acc@0.5 on ScanRefer~\cite{chen2020scanrefer}, and Acc@0.25 on Nr3D~\cite{achlioptas2020referit3d}. For situation estimation, we evaluate localization accuracy using Acc@0.5m and Acc@1.0m, and orientation accuracy using Acc@15° and Acc@30°. For 3D question answering, we report exact match accuracy (EM) and the refined exact match protocol (EM-R) proposed by LEO~\cite{huang2024embodied} on SQA3D~\cite{ma2022sqa3d}. Similarly, we report CIDEr, BLEU-4, METEOR, ROUGE, and EM on ScanQA~\cite{azuma2022scanqa}. For 3D-assisted dialog and task decomposition, we adopt BLEU-1, BLEU-4, METEOR, and ROUGE on 3D-LLM held-in dataset~\cite{hong20233d}.

\noindent \textbf{Implementation Details.}
We adopt Qwen2.5-VL-72B~\cite{bai2025qwen2} as Planning Agent and GPT-4o (gpt-4o-2024-08-06~\cite{hurst2024gpt}) as Perception Agent. The rendered images are generated at a resolution of $512 \times 512$ pixels. We set the viewpoint matching threshold to $\tau_D = 0.8$. The number of planned viewpoints per interaction round is fixed at $K=8$. The default number of Perception Agent is set to 1. The number of interaction rounds is limited to a maximum of 6. Following SeeGround~\cite{li2025seeground}, we exclude the top 0.3m of the scene to better account for closed room setup. Additional details are provided in \textcolor{cvprblue}{Appendix A.2}.

\subsection{Quantitative Results}
\label{sec:4.2}

\begin{table*}[t]
    \centering
    \caption{{Evaluation of 3D question answering} on SQA3D~\cite{ma2022sqa3d} test set and ScanQA~\cite{azuma2022scanqa} validation set. \textcolor{gray}{Gray font} means the refined exact match, EM-R.}
    \resizebox{1.0\linewidth}{!}{
    \begin{tabular}{lcccccccccccc}
        \toprule
        \multirow{2}{*}{{Method}} & \multicolumn{7}{c}{{SQA3D}} & \multicolumn{5}{c}{{ScanQA}} \\
        \cmidrule(lr){2-8} \cmidrule(lr){9-13}
         & What & Is & How & Can & Which & Other & Overall  & CIDEr & BLEU-4 & METEOR & ROUGE & EM \\
    \midrule\midrule
        \multicolumn{13}{l}{\textit{Supervision: Fully Supervised}}\\
        ScanQA~\cite{azuma2022scanqa}           & 28.6 & 65.0 & 47.3 & 66.3 & 43.9 & 42.9 & 45.3 & 60.2 & 10.8 & 12.6 & 31.1 & 20.9 \\
        SQA3D~\cite{ma2022sqa3d}                & 33.5 & 66.1 & 42.4 & 69.5 & 43.0 & 46.4 & 47.2 & -    & -    & -    & -    & -    \\
        3D-VLP~\cite{jin2023context} & - & - & - & - & - & - & - & 67.0 & - & 13.5 & 34.5 & 21.7 \\
        3D-LLM~\cite{hong20233d}                & 36.5 & 65.6 & 47.2 & 68.8 & 48.0 & 46.3 & 48.1 & 69.4 & 12.0 & 14.5 & 35.7 & 20.5 \\
        LEO~\cite{huang2024embodied}            & -    & -    & -    & -    & -    & -    & 50.0 \textcolor{gray}{(52.4)} & 101.4 & 13.2 & 20.0 & 49.2 & 24.5 \\
        SIG3D~\cite{man2024situational}         & 35.6 & 67.2 & 48.5 & 71.4 & 49.1 & 45.8 & 52.6 & 68.8 & 12.4 & 13.4 & 35.9 & -  \\
        Chat-Scene~\cite{huang2024chat}         & 45.4 & 67.0 & 52.0 & 69.5 & 49.9 & 55.0 & 54.6 & 87.7 & 14.3 & 18.0 & 41.6 & 21.6 \\
        LSceneLLM~\cite{zhi2025lscenellm}            & - & - & - & - & - & - & - & 88.2 & - & 18.0 & 40.8 & - \\
        Scene-LLM~\cite{fu2025scene}            & 40.9 & 69.1 & 45.0 & 70.8 & 47.2 & 52.3 & 54.2 & 80.0 & 12.0 & 16.6 & 40.0 & 27.2 \\
        Video-3D LLM~\cite{zheng2025video}      & 51.1 & 72.4 & 55.5 & 69.8  & 51.3  & 56.0    & 58.6 & 102.1 & 16.3 & 19.8 & 49.0 & 30.1 \\
        3DRS~\cite{huang2025mllms}            & 54.4 & 75.2 & 57.0 & 72.2& 49.9 & 59.0 & 60.6  & 104.8 & 17.2 & 20.5 & 49.8 & 30.3 \\
        Ross3D~\cite{wang2025ross3d}            & 56.0 & 79.8 & 60.6 & 70.4& 55.3 & 60.1 & 63.0 \textcolor{gray}{(65.7)} & 107.0 & 17.9 & 20.9 & 50.7 & 30.8 \\
    \midrule
    \multicolumn{13}{l}{\textit{Supervision: Zero-shot}}\\
        Agent3D-Zero~\cite{zhang2024agent3d}               & - & - & - & - & - & - & - & 71.8  & 4.4 & 16.0 & 37.0 & 17.5  \\
        SpatialPrompting~\cite{taguchi2025spatialprompting}               & 48.7 & 64.3 & \textbf{53.6} & 58.9 & 33.6 & 55.3 & 52.7 & 87.7  & 10.9 & 16.9 & 43.4 & 27.3  \\
        \rowcolor{resultcolor}
        {Ours}                    & \textbf{50.2} & \textbf{65.0} & 46.7 & \textbf{67.8} & \textbf{46.4} & \textbf{55.7} & \textbf{54.8 \textcolor{gray}{(57.1)}} & \textbf{91.1}    & \textbf{12.3}    & \textbf{17.4}    & \textbf{44.5}    & \textbf{28.7}  \\
        \bottomrule
    \end{tabular}}
    \label{tab:sqa3d_QA}
\end{table*}
\begin{figure}[t!]
    \centering
        \begin{minipage}[t]{.47\linewidth}
            \centering
            \captionof{table}{Evaluation of situation estimation on SQA3D~\cite{ma2022sqa3d} test set. {``Separate''} indicates models trained exclusively for situation estimation. \textit{``*''} denotes experimental results reported in ~\cite{yuan2025empowering}.}
            \label{tab:sqa3d_situation}
            \resizebox{\linewidth}{!}{
            \begin{tabular}{lcccc}
                \toprule 
                \multirow{2}{*}{{Method}} & \multicolumn{2}{c}{{Localization}} & \multicolumn{2}{c}{{Orientation}} \\
                \cmidrule(lr){2-3} \cmidrule(lr){4-5}
                 & Acc@0.5m & Acc@1.0m & Acc@15$^\circ$ & Acc@30$^\circ$ \\
                \midrule\midrule
                \multicolumn{5}{l}{\textit{Supervision: Fully Supervised}} 
                \\
                SQA3D~\cite{ma2022sqa3d} & 9.5  & 29.6 & 8.7  & 16.5 
                \\
                SQA3D ({separate})& 10.3 & 31.4 & 17.1 & 22.8 
                \\
                3D-VisTA~\cite{zhu20233d}& 11.7 & 34.5 & 16.9 & 24.2 
                \\
                SIG3D$^{*}$~\cite{man2024situational}  & 16.8 & 35.2 & 23.4 & 26.3 
                \\
                3DSA-LLM~\cite{yuan2025empowering}  & 17.4 & 36.9 & 24.1 & 28.5 
                \\
                \midrule
                \multicolumn{5}{l}{\textit{Supervision: Zero-shot}} \\
                \rowcolor{resultcolor}
                {Ours}   & \textbf{20.7} & \textbf{48.9} & {22.8} & \textbf{32.2} \\
                \bottomrule
            \end{tabular}}
        \end{minipage}\hspace{0.04\linewidth}%
        \begin{minipage}[t]{.49\linewidth}
            \centering
            \captionof{table}{Evaluation of 3D-assisted Dialog and task decomposition on 3D-LLM Held-In Dataset~\cite{hong20233d}. B-1, B-4, M, and R denote BLEU-1, BLEU-4, METEOR, and ROUGE, respectively.}
            \label{tab:held-in}
            \resizebox{\linewidth}{!}{
                \begin{tabular}{lcccccccc}
                \toprule
                \multirow{2}{*}{Method} & \multicolumn{4}{c}{3D-assisted Dialog} & \multicolumn{4}{c}{Task Decomposition} \\
                \cmidrule(lr){2-5} \cmidrule(lr){6-9}
                 & B-1 & B-4 & M & R & B-1 & B-4 & M & R \\
                \midrule\midrule
                \multicolumn{9}{l}{\textit{Supervision: Fully Supervised}} \\
                flant5~\cite{chung2024scaling} & 27.4 & 8.7 & 9.5 & 27.5 & 25.5 & 6.0 & 13.9 & 28.4 \\
                flamingo~\cite{alayrac2022flamingo} & 30.6 & 9.1 & 10.4 & 27.9 & 33.1 & 7.3 & 16.1 & 33.2 \\
                BLIP2-flant5~\cite{li2023blip} & 32.4 & 9.5 & 11.0 & 29.5 & 33.1 & 6.9 & 15.5 & 34.0 \\
                3D-LLM~\cite{hong20233d}  & 39.0 & 16.6 & 18.9 & 39.3 & 33.9 & 7.4 & 15.9 & 37.8 \\
                \midrule
                \multicolumn{9}{l}{\textit{Supervision: Zero-shot}} \\
                Agent3D-Zero~\cite{zhang2024agent3d}  & 32.8 & 9.8 & 19.3 & 39.3 & 42.0 & 15.5 & \textbf{22.9} & 45.1 \\
                \rowcolor{resultcolor}
                {Ours}  & \textbf{47.4} & \textbf{24.0} & \textbf{28.6} & \textbf{47.3} & \textbf{43.8} & \textbf{24.9} & 21.0 & \textbf{48.8} \\
                \bottomrule
                \end{tabular}
            }
        \end{minipage}
\end{figure}

\noindent \textbf{3D Question Answering.} 
As illustrated in \cref{tab:sqa3d_QA}, the quantitative comparison shows that our method outperforms existing zero-shot methods in most evaluation metrics. Compared with SpatialPrompting~\cite{taguchi2025spatialprompting}, it achieves improvements of 2.1\% EM on SQA3D~\cite{ma2022sqa3d} and 3.4 CIDEr on ScanQA~\cite{azuma2022scanqa}. Moreover, our method already exceeds most fully supervised approaches, such as LEO~\cite{huang2024embodied}, Chat-Scene~\cite{huang2024chat}, and Scene-LLM~\cite{fu2025scene}. These results demonstrate that our approach attains a better understanding of spatial layouts and fine-grained object semantics by explicitly documenting the 3D scene.

\noindent \textbf{Situation Estimation.}
In \cref{tab:sqa3d_situation}, the results on SQA3D~\cite{ma2022sqa3d} demonstrate that our method achieves the best localization performance over existing methods, with 20.7\% Acc@0.5m and 48.9\% Acc@1.0m. Owing to its flexible viewpoint planning, our method can effectively map situation descriptions to accurate spatial locations and orientations, thereby enhancing spatial understanding.

\noindent \textbf{3D-assisted Dialog.} 
The results in \cref{tab:held-in} illustrate that our method establishes a new state-of-the-art zero-shot performance on 3D-LLM Held-In dataset \cite{hong20233d}, surpassing fully supervised method 3D-LLM~\cite{hong20233d} and zero-shot method Agent3D-Zero~\cite{zhang2024agent3d} by 8.4 and 14.6 BLEU-1, respectively.

\begin{table*}[t]
    \centering
    \caption{Evaluation of 3DVG on ScanRefer~\cite{chen2020scanrefer} and Nr3D~\cite{achlioptas2020referit3d} validation sets. In ScanRefer, queries are annotated as {``Unique''} (single target) or {``Multiple''} (with same-class distractors). In Nr3D, queries are categorized as ``Easy'' (one distractor) or {``Hard''} (multiple distractors), and as {``View-Dependent''} or {``View-Independent''}, depending on whether grounding requires specific viewpoints. \textit{0.25} and \textit{0.5} denote Acc@0.25 and Acc@0.5, respectively. We also follow VLM-Grounder~\cite{xu2024vlm} to adopt the same \textit{``250 queries''} setting for a fair comparison.}
    
    \resizebox{\linewidth}{!}{
    \begin{tabular}{p{3.0cm} *{6}{>{\centering\arraybackslash}p{1.1cm}}*{5}{>{\centering\arraybackslash}p{1.1cm}}}
        \toprule 
        \multirow{3}{*}{{Method}} & \multicolumn{6}{c}{{ScanRefer}}& \multicolumn{5}{c}{{Nr3D}}\\
        \cmidrule(lr){2-7} \cmidrule(lr){8-12}
        & \multicolumn{2}{c}{{Unique}} & \multicolumn{2}{c}{{Multiple}} & \multicolumn{2}{c}{{Overall}} & {Easy} & {Hard} & {Dep.} & {Indep.} & {Overall}
        \\
         & \textit{0.25} & \textit{0.5} & \textit{0.25} & \textit{0.5} & \textit{0.25} & \textit{0.5} & \textit{0.25} & \textit{0.25} & \textit{0.25} & \textit{0.25} & \textit{0.25}
        \\
        \midrule\midrule
        \multicolumn{12}{l}{\textit{Supervision: Fully Supervised}} \\
        ScanRefer~\cite{chen2020scanrefer}  & 67.6 & 46.2 & 32.1 & 21.3 & 39.0 & 26.1 & - & - & - & - & -
        \\
        ReferIt3DNet~\cite{achlioptas2020referit3d} & - & - & - & - & - & - & 43.6 & 27.9 & 32.5 & 37.1 & 35.6
        \\
        3DVG-T~\cite{zhao20213dvg}  & 77.2 & 58.5 & 38.4 & 28.7 & 45.9 & 34.5 & 48.5 & 34.8 & 34.8 & 43.7 & 40.8
        \\
        BUTD-DETR~\cite{jain2022bottom}  & 84.2 & 66.3 & 46.6 & 35.1 & 52.2 & 39.8 & 60.7 & 48.4 & 46.0 & 58.0 & 54.6
        \\
        3D-VisTA~\cite{zhu20233d}  & 81.6 & 75.1 & 43.7 & 39.1  & 50.6 & 45.8 & 72.1 & 56.7 & 61.5 & 65.1 & 64.2
        \\
        MiKASA~\cite{chang2024mikasa} & - & - & - & -  & -  & - & 69.7 & 59.4 & 65.4 & 64.0 & 64.4
        \\
        Video-3D LLM~\cite{zheng2025video} & 88.0 & 78.3 & 50.9 & 45.3  & 58.1 & 51.7  & - & - & - & - & -
        \\
        3DRS~\cite{huang2025mllms} & 87.4 & 77.9 & 57.0 & 50.8  & 62.9 & 56.1 & - & - & - & - & -
        \\
        Ross3D~\cite{wang2025ross3d} & 87.2 & 77.4 & 54.8 & 48.9  & 61.1 & 54.4 & - & - & - & - & -
        \\
        \midrule
        \multicolumn{12}{l}{\textit{Supervision: Zero-shot (250 queries)}} \\
        {VLM-Grounder}~\cite{xu2024vlm}  &  {66.0} &  {29.8} &  {48.3} &  {33.5} &  {51.6} &  {32.8} & 55.2 & 39.5 & 45.8 & 49.4 & 48.0
        \\
        {SeqVLM}~\cite{lin2025seqvlm}  &  {77.3} &  {72.7} &  {47.8} &  {41.3} &  {55.6} &  {49.6} & 58.1 & 47.4 & 51.0 & 54.5 & 53.2
        \\
        {SPAZER}~\cite{jin2025spazer}  &  {80.9} &  {72.3} &  {51.7} &  {43.4} &  {57.2} &  {48.8} & 68.0 & 58.8 & 59.9 & 66.2 & 63.8
        \\
        \rowcolor{resultcolor}
        {Ours} & \textbf{83.0} & \textbf{74.5} & \textbf{52.2} & \textbf{44.3} & \textbf{58.0} & \textbf{50.0} & \textbf{69.9} & \textbf{60.5} & \textbf{61.5} & \textbf{68.2} & \textbf{65.6}
        \\
        \midrule
        \multicolumn{12}{l}{\textit{Supervision: Zero-shot (full)}} \\
        LLM-Grounder~\cite{yang2024llm}  & - & -  & - & - & 17.1 & 5.3 & - & -  & - & - & -
        \\
        ZSVG3D~\cite{yuan2024visual}  & 63.8 & 58.4 & 27.7 & 24.6 & 36.4 & 32.7 & 46.5 & 31.7 & 36.8 & 40.0 & 39.0
        \\
        SeeGround~\cite{li2025seeground}  & {75.7} & {68.9} & 34.0 & {30.0} & 44.1 & 39.4 & 54.5 & 38.3 & 42.3 & 48.2 & 46.1
        \\
        CSVG~\cite{yuan2024solving}  & 68.8 & 61.2 & {38.4} & 27.3 & {49.6} & {39.8} & \textbf{67.1} & {51.3} & {53.0} & {62.5} & {59.2}
        \\
        \rowcolor{resultcolor}
        {Ours} & \textbf{81.2} & \textbf{72.9} & \textbf{50.8} & \textbf{43.9} & \textbf{58.1} & \textbf{50.9} & {65.7} & \textbf{57.0} & \textbf{56.7} & \textbf{63.7} & \textbf{61.2}
        \\
        \bottomrule
    \end{tabular}}
    \label{tab:3DVG}
\end{table*}

\noindent \textbf{Task Decomposition.} 
As demonstrated in \cref{tab:held-in}, our method outperforms existing zero-shot methods in most evaluation metrics. Compared with zero-shot method Agent3D-Zero~\cite{zhang2024agent3d}, our method achieves improvements of 1.8 BLEU-1, 9.4 BLEU-4, and 3.7 ROUGE on the 3D-LLM Held-In dataset~\cite{hong20233d}.

\noindent \textbf{3D Visual Grounding.}
As shown in \cref{tab:3DVG}, compared with the previous state-of-the-art zero-shot method CSVG~\cite{yuan2024solving}, our method achieves substantial improvements of 8.5\% Acc@0.25 and 11.1\% Acc@0.5 on ScanRefer~\cite{chen2020scanrefer}, and 2.0\% Acc@0.25 on Nr3D~\cite{achlioptas2020referit3d}. Moreover, even on the subset of 250 queries proposed by the VLM-Grounder~\cite{xu2024vlm}, our method still achieves advanced performance. Notably, it also achieves competitive performance on the challenging ``Multiple'' subset, indicating its ability to efficiently disambiguate among multiple similar objects through unified spatial cognition.

\begin{figure}[t!]
    \centering
    \begin{minipage}[]{0.55\linewidth}
        \centering
        \includegraphics[width=\linewidth]{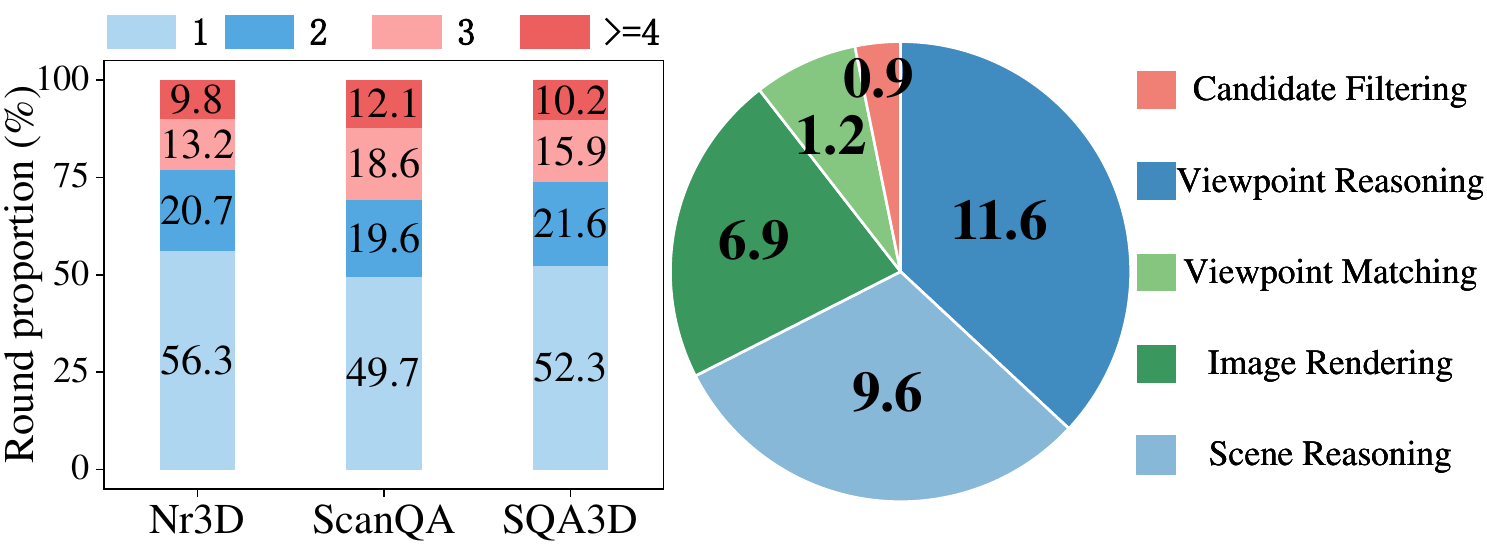} 
        \captionof{figure}{Left: Distribution of interaction rounds. Right: Inference time of each step.}
        \label{fig:round}
    \end{minipage}
    \begin{minipage}[]{0.44\linewidth}
        \centering
        \captionof{table}{Inference efficiency comparison with zero-shot methods. ``*'' denotes our reproduced results.}
        \label{tab:token}
        \resizebox{\linewidth}{!}{
            \begin{tabular}{lccc}
                \toprule
                {Method} & {Tokens} & {Time (s)} & {Acc.} \\
            \midrule\midrule

                VLM-Grounder  & - & 50.3 & 48.0 \\
                SeqVLM*        & 7.5k & 42.3 & 53.2 \\
                \rowcolor{resultcolor}
                Ours   & \textbf{4.3k}  & \textbf{30.2} & \textbf{65.6} \\
                \bottomrule
            \end{tabular}}
    \end{minipage}
\end{figure}

\noindent\textbf{Interaction Rounds and Inference Efficiency.}
As shown in the left of \cref{fig:round}, we investigate the distribution of interaction rounds across different benchmarks. We observe that our multi-agent collaborative framework efficiently resolves approximately 50\% of the queries in a single round, while only a small fraction requires four or more iterations. Meanwhile, Nr3D~\cite{achlioptas2020referit3d} exhibits fewer interaction rounds compared to other benchmarks. This is because Nr3D primarily involves selecting and localizing the target object, whereas ScanQA~\cite{azuma2022scanqa} and SQA3D~\cite{ma2022sqa3d} additionally require understanding more complex object semantics, spatial relationships, and intricate layouts. In the right of \cref{fig:round}, we break down the inference time of each step within a single round, which in total takes 30.2s. The reasoning processes of the two agents take up a significant portion of the total time. For the multi-round case, the total inference time increases linearly with the number of interaction rounds. Moreover, \cref{tab:token} shows that our method significantly reduces the token cost from 7.5k to 4.3k and achieves higher efficiency from 41.3s to 30.2s, while still achieving strong performance on Nr3D~\cite{achlioptas2020referit3d}. This is because our method leverages the cognitive map to filter out a large number of erroneous candidates and plans viewpoints around the critical target, which avoids the need to sample and filter out all video frames.

\begin{figure}[t!]
    \centering
    \begin{minipage}[]{0.52\linewidth}
        \centering
        \captionof{table}{Ablation on different components.}
        \label{tab:ablation}
        \resizebox{\linewidth}{!}{
            \begin{tabular}{lccc}
            \toprule
            {Method} & {Nr3D} & {SQA3D} & {ScanQA} \\
            \midrule\midrule
            \rowcolor{resultcolor}
            {Ours (Full)}                      & \textbf{56.7}  & \textbf{53.2} & \textbf{26.1} \\
            {w/o Holistic Cognitive Map}        & {47.8} & {49.3} & {18.6}  \\
            {w/o Filtering and Feedback}        & {49.3} & {50.4} & {20.9} \\
            {w/o Rendered Image}                & {52.7} & {50.6} & {21.0} \\
            {w/o Real Image}                    & {53.9} & {51.7} & {21.4} \\
            \bottomrule
            \end{tabular}}
    \end{minipage}
    \begin{minipage}[]{0.46\linewidth}
        \centering
        \captionof{table}{Performance comparison with different baseline.}
        \label{tab:baseline}
        \resizebox{\linewidth}{!}{
        \begin{tabular}{lccc}
        \toprule
        {Method} & {Nr3D} & {SQA3D} & {ScanQA} \\
        \midrule\midrule
            {Random}                  & {13.8}  & {12.6} & {6.4} \\
            {Co-Visible Template}        & {43.1} & {42.0} & {13.6}  \\
            \rowcolor{resultcolor}
            Ours       & \textbf{56.7} & \textbf{53.2} & \textbf{26.1} \\
            \bottomrule
        \end{tabular}}
    \end{minipage}
\end{figure}

More experimental results under the multiple Perception Agents setting are provided in \textcolor{cvprblue}{Appendix B.1}.

\subsection{Ablation Study}
\label{sec:4.3}

In this section, we adopt Acc@0.25 as the default evaluation metric for Nr3D~\cite{achlioptas2020referit3d}, and EM for SQA3D~\cite{ma2022sqa3d} and ScanQA~\cite{azuma2022scanqa}. We employ Qwen2.5-VL-72B~\cite{bai2025qwen2} as both Planning Agent and Perception Agent.

\noindent\textbf{Effect of Holistic Cognitive Map.}
As illustrated in \cref{tab:ablation}, removing the object information table and retaining only the BEV leads to notable performance drops of 8.9\% Acc@0.25 on Nr3D~\cite{achlioptas2020referit3d} and 3.9\% EM on SQA3D~\cite{ma2022sqa3d}. We find that without the holistic cognitive map to refine the cognition of 3D scenes, this variant is forced to repeatedly process previously observed objects, leading to redundant planning and reduced efficiency.

\noindent\textbf{Effect of Filtering and Feedback.}
We disable both candidate filtering and feedback, which jointly sustain the collaborative process between the two agents. As demonstrated in \cref{tab:ablation}, this variant results in a notable performance degradation of 7.4\% Acc@0.25 on Nr3D~\cite{achlioptas2020referit3d}. Without candidate filtering and feedback, mismatched objects remain in the candidate map, thereby preventing Planning Agent from focusing on critical objects.

\noindent\textbf{Effect of Image Sources.}
Our method balances visual fidelity and perspective coverage by prioritizing real images and using rendered images only when necessary. As shown in \cref{tab:ablation}, without Rendered Images results in a performance drop of 5.1\% EM on ScanQA~\cite{azuma2022scanqa}. This performance gap occurs because, without rendered images, this variant can only perceive the 3D scene from limited and fixed viewpoints, thereby missing critical perspectives that are crucial for spatial localization. In contrast, w/o Real Images leads to a performance drop of 4.7\% EM on ScanQA~\cite{azuma2022scanqa}. This variant limits the ability of our method to capture fine-grained object semantics, such as material and texture details. These results highlight the complementary roles of rendered and real images in supporting a comprehensive understanding of the 3D scene.


\begin{wraptable}{r}{0.3\linewidth}
    \centering
    \caption{{Ablation study} of the MLLM on Nr3D~\cite{achlioptas2020referit3d}}
    \resizebox{\linewidth}{!}{
    \begin{tabular}{l|c}
        \toprule
        {Method} & {Overall}\\
    \midrule\midrule
        {Qwen2.5-VL-7B}                      & 49.3   \\

        {Qwen2-VL-72B}        & {54.2}   \\
        {Qwen2.5-VL-72B}        & {56.7} \\
        \midrule
        \rowcolor{resultcolor}
        {GPT-4o}                & \textbf{61.2}\\
        \bottomrule
    \end{tabular}}
    \label{tab:MLLM}

\end{wraptable}

\noindent\textbf{Influence of MLLMs.}
The MLLM serves as the "brain" of our method, effectively parsing the query and performing spatial relationship reasoning. To further study how the choice of the MLLM affects overall performance, we perform a comparative analysis of several models under the same evaluation setting, including open-source backbones (Qwen2.5-VL-72B~\cite{bai2025qwen2}, Qwen2.5-VL-7B~\cite{bai2025qwen2}, and Qwen2-VL-72B~\cite{qwen2vl}) and a closed-source alternative (GPT-4o~\cite{hurst2024gpt}). As demonstrated in \cref{tab:MLLM}, we observe that GPT-4o exhibits stronger 3D scene understanding capabilities than open-source models. Moreover, our method still outperforms most zero-shot methods even when using smaller backbones, indicating that its effectiveness stems from explicit scene cognition, candidate filtering, and comprehensive observation, rather than relying solely on more advanced MLLMs.


\subsection{Discussion}
\label{sec:4.4}
\noindent\textbf{Effectiveness of Viewpoint Planning.}
As mentioned in \cref{sec:3.2}, our method actively plans observation perspectives to capture scene information. To analyze the effectiveness of such design, we further compare the following two baselines: (1) Random, which randomly plans viewpoints based on the BEV~\cite{zhang2024agent3d}. (2) Co-Visible Template, which generates fixed near and far viewpoints around candidate objects to ensure broad scene coverage. As demonstrated in \cref{tab:baseline}, Random performs poorly because it may fail to reliably capture the target object. Co-Visible Template performs better than Random by increasing scene coverage. In contrast, our method leverages the cognitive map to directly observe the 3D scene and actively capture more informative viewpoints, leading to performance improvements of 42.9\% and 13.6\% Acc@0.25 over Random and Co-Visible Template on Nr3D~\cite{achlioptas2020referit3d}, respectively. These results reveal an important principle: the key to 3D understanding lies in capturing decisive visual evidence from critical perspectives, rather than increasing coverage alone.

\begin{figure}[t]
\centering
\includegraphics[width=0.9\linewidth]{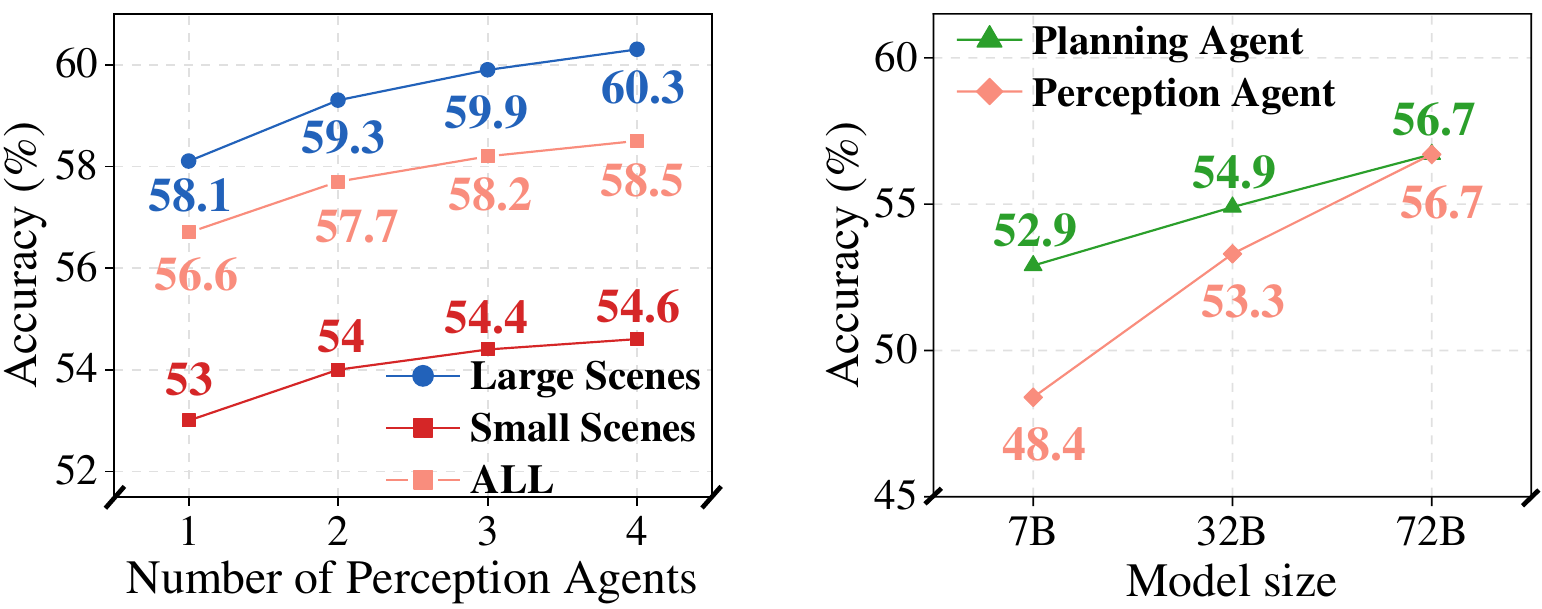} 
\caption{Left: Influence of the number of Perception Agents on Nr3D~\cite{achlioptas2020referit3d}. Right: Influence of the model size of Planning Agent and Perception Agent on Nr3D~\cite{achlioptas2020referit3d}.
}
\label{fig:insight}
\end{figure}

\noindent\textbf{More Perception Agents across Scene Scales.}
As demonstrated in the left of \cref{fig:insight}, we analyze the effect of varying the number of Perception Agents across different scene scales, with each Perception Agent tasked with perceiving a different region. For scene scale analysis, we divide ScanNet~\cite{dai2017scannet} into small scenes (average area 16.32m$^2$) and large scenes (average area 40.85m$^2$). We find that as the number of Perception Agents increases, the performance continues to improve, with larger improvements observed in large scenes compared to small ones. However, the rate of improvement gradually slows down. This is because the relatively small scene scale of ScanNet~\cite{dai2017scannet} limits the advantage of our multi-agent cognition, which is better suited for larger-scale 3D scenes. 

\noindent\textbf{Better Input or Better Perception?}
Planning Agent provides query-relevant inputs, and Perception Agent is tasked to perform scene perception. This design motivates an investigation centered around the question: \textit{Which matters more, better input or better perception?} In the right of \cref{fig:insight}, we further conduct two symmetric experiments by keeping one agent at 72B while varying the other agent's model size. We find that better input is a promising direction that provides reliable support for subsequent perception, while better perception further enhances the understanding of the 3D scene and leads to greater improvement. Moreover, the effectiveness of both improvements is largely influenced by model size, and continued advancements in MLLMs may further unlock its potential.


For a detailed sensitivity analysis of the viewpoint matching threshold $\tau_D$ and the number of planned viewpoints $K$, please refer to \textcolor{cvprblue}{Appendix B.2}.

\subsection{Qualitative Results}
\label{sec:4.5}

\begin{figure*}[t]
    \centering
    \subcaptionbox{3D visual grounding
        \label{fig:visual_vg}
    }{%
        \includegraphics[height=0.159\textheight]{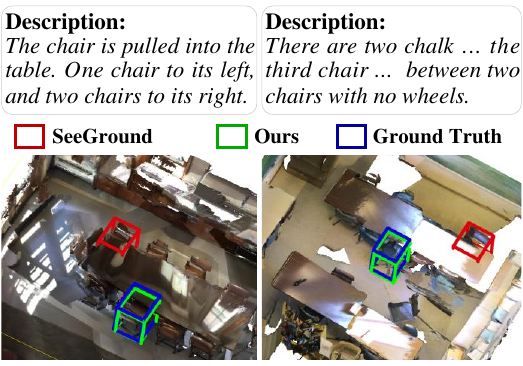}
    }%
    \raisebox{0pt}[0pt][0pt]{%
      \begin{tikzpicture}
        \draw[dash pattern=on 2pt off 1.2pt, line width=0.6pt] 
          (0,0.069\textheight) -- (0,-0.069\textheight);
      \end{tikzpicture}%
    }%
    \subcaptionbox{Situated understanding
        \label{fig:visual_sqa}
    }{%
        \includegraphics[height=0.159\textheight]{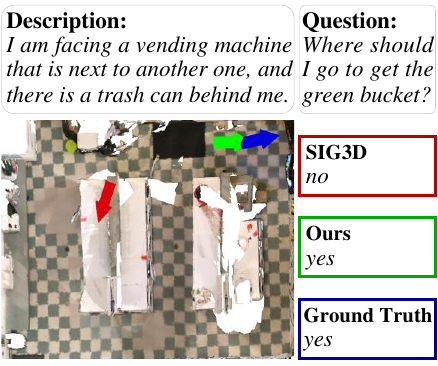}
    }%
    \raisebox{0pt}[0pt][0pt]{%
      \begin{tikzpicture}
        \draw[dash pattern=on 2pt off 1.2pt, line width=0.6pt] 
          (0,0.069\textheight) -- (0,-0.069\textheight);
      \end{tikzpicture}%
    }%
    \subcaptionbox{Visual question answering
        \label{fig:visual_scanqa}
    }{%
        \includegraphics[height=0.159\textheight]{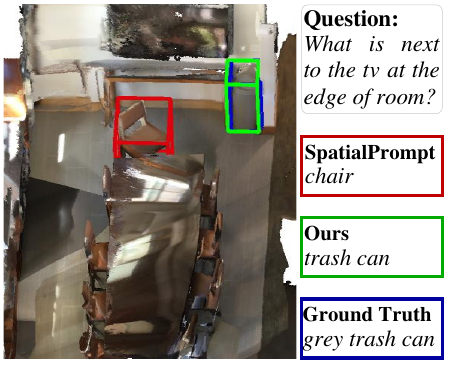}
    }%
    \caption{
    Qualitative comparison of our method with previous approaches across various 3D understanding tasks on ScanRefer~\cite{chen2020scanrefer}, SQA3D~\cite{ma2022sqa3d}, and ScanQA~\cite{azuma2022scanqa}, respectively.
    }
    \label{fig:visualize}
\end{figure*}

\cref{fig:visualize} illustrates the qualitative results on ScanRefer~\cite{chen2020scanrefer}, SQA3D~\cite{ma2022sqa3d}, and ScanQA~\cite{azuma2022scanqa}. In \cref{fig:visual_vg}, multiple identical objects (chairs) appear in the scene. Our method can accurately understand spatial relationships (\textit{e.g.}, ``to its right'', ``between'') to localize the target object, but SeeGround~\cite{li2025seeground} relies on single-view localization, which limits its ability to resolve spatial ambiguity. In \cref{fig:visual_sqa}, the query includes egocentric situations (\textit{e.g.}, ``facing a vending machine'', ``behind me'') and fine-grained semantics (\textit{e.g.}, ``green bucket''). Our explicit cognitive map enables the model to directly observe the 3D scene and understand egocentric spatial locations, and it further verify the orientation through flexible viewpoint observations. In contrast, SIG3D relies on fixed anchors and struggles to interpret complex location descriptions. In \cref{fig:visual_scanqa}, SpatialPrompting~\cite{taguchi2025spatialprompting} fails to capture the spatial layout (\textit{e.g.}, ``edge of room'') because its object-centric keyframe selection ignores relations among objects and scene layout.



\section{Conclusion}
\label{sec:conclusion}
In this paper, we propose a collaborative multi-agent framework for zero-shot 3D scene understanding. Unlike previous methods that rely on static object-centric keyframe selection from the scene video, we integrate a Planning Agent for flexible viewpoint planning and critical perspective completion, and a Perception Agent for explicit environment perception, enabling more comprehensive cognition of query-relevant scene information. Extensive experiments demonstrate that our approach achieves promising performance across a variety of 3D tasks, highlighting its potential as a foundation for future multi-agent systems in 3D scene understanding.

\noindent\textbf{Limitations and Future Work.}
We follow previous works and adopt a pre-trained 3D detection model to obtain 3D bounding boxes, making overall performance related to their detection quality. In future work, we will integrate a high-fidelity online scene reconstruction model to refine detection errors and reduce this dependency. For more detailed discussions of limitations, failure cases, and error types, please refer to the Appendix.


\section*{Acknowledgment}
This work is supported by the National Natural Science Foundation of China (U23B2013, U2441242 and 62276176). This work was also partly supported by the SICHUAN Provincial Natural Science Foundation (No. 2024NSFJQ0023).

%
%
\clearpage
\bibliographystyle{splncs04}
\bibliography{main}

\newpage
\appendix
\setcounter{page}{1}
\begin{center}
    {\Large \textbf{Agentic Collaborative Cognition for Zero-Shot 3D Understanding}}\\
    \vspace{5mm}
    {\large Supplementary Materials}
\end{center}

\section{Implementation Details}
\label{secs_supp:implement}
In this section, we present additional implementation details, including the overall algorithm, model details, and agent-specific prompts.

\subsection{Overall Algorithm}
The overall algorithm is summarized in \cref{alg:framework}. The default setting contains a Planning Agent and a Perception Agent. Planning Agent analyzes cognitive map to retrieve candidates, plans viewpoints and maps them to obtain images. Perception Agent documents object attributes, matches current observations with the query, and provides feedback to refine candidates or generates answer.

\subsection{Model Details}
For all MLLMs, we set the temperature to 0.2 to reduce response randomness to improve the reproducibility of the results. All experiments are conducted on 4 NVIDIA H100 GPUs. For 3D bounding boxes and category labels, we follow SPAZER~\cite{jin2025spazer} and use the Mask3D~\cite{schult2023mask3d} detector for ScanRefer~\cite{chen2020scanrefer} and 3D-LLM held-in datasets~\cite{hong20233d}. For Nr3D~\cite{achlioptas2020referit3d}, SQA~\cite{ma2022sqa3d}, and ScanQA~\cite{azuma2022scanqa}, we follow the default setting of Nr3D~\cite{achlioptas2020referit3d} and adopt the instance annotations from ScanNet~\cite{dai2017scannet}.

\subsection{Prompts Design}
Our method assigns each MLLM a specialized prompt. Specifically, we first identify key objects using the prompt in \cref{tab:supp_key_object}, which is similar with SeeGround~\cite{li2025seeground}. The prompt in \cref{tab:supp_planning_agent} guides \textbf{Planning Agent} to perform viewpoint planning, while the prompt in \cref{tab:supp_perception_agent} guides \textbf{Perception Agent} to document the 3D scene and generate the final answer. The prompts in \cref{tab:supp_planning_agent} and \cref{tab:supp_perception_agent} are illustrated with the 3D Visual Grounding task as a representative example. 

\section{Additional Results and Analysis}
\label{secs_supp:additional_results_analysis}
In this section, we provide additional experimental results, including evaluations under the multiple Perception Agents setting, sensitivity analysis of hyperparameters, and a full comparison with existing methods. Moreover, we provide additional qualitative results to illustrate the effectiveness of our method.

\begin{algorithm}[H]
\caption{Collaborative Multi-Agent Framework}
\label{alg:framework}
\begin{algorithmic}[1]
\STATE \textbf{Input:} Query $Q$, 3D point cloud $\mathcal{P}$, real image set $\mathcal{I} = \{I_m\}_{m=1}^{M}$ with rotation $R_m$ and translation $T_m$, viewpoint matching threshold $\tau_D $, maximum interaction rounds $Round_{\max}$
\STATE \textbf{Output:} Final answer $y$

\STATE Build initial holistic cognitive map $\mathcal{M} = \{\mathcal{T}, I_B\}$ by 3D detection on $\mathcal{P}$
\STATE Initialize $a_i \leftarrow \emptyset$ for all $\sigma_i = \{b_i,l_i,a_i\} \in \mathcal{T}$
\STATE Parse key objects $\mathcal{O} = \{ o_j \}_{j=1}^{J}$ from $Q$
\STATE $t \leftarrow 0$, $\mathcal{F}^0 \leftarrow \emptyset$, status $\leftarrow$ \textit{Continue}

\WHILE{status = \textit{Continue} \AND $t < Round_{\max}$}
    \STATE \textit{// Planning Agent}
    \IF{$t = 0$}
        \STATE $\mathcal{C}^t \leftarrow \{\sigma_i \in \mathcal{T} \mid \mathrm{Match}(\sigma_i, o_j)=1\}$
    \ELSE
        \STATE $\mathcal{C}^t \leftarrow$ refined candidates from $\mathcal{C}^{t-1}$ based on feedback $\mathcal{F}^t$

    \ENDIF
    \STATE Plan viewpoints $\mathcal{V}^t = \{V_k^{t}\}_{k=1}^{K}$ based on $\mathcal{C}^t$

    
    \FOR{$k = 1$ to $K$}
        \STATE Compute pose distances $D_{k,m}^{t}$ between $\mathcal{V}_k^t$ and $I_{m}$
        \STATE $\hat{m} \leftarrow \mathop{\arg\min}\limits_{m=1,\ldots,M}\ D_{k,m}^{t}$
        \IF{$D_{k,\hat{m}}^{t} < \tau_D$}
            \STATE $I_k^{t} \leftarrow I_{\hat{m}}$
        \ELSE
            \STATE $I_k^{t} \leftarrow$ a render image based on $\mathcal{P}$
        \ENDIF
    \ENDFOR
    \STATE $\mathcal{I}^{t} \leftarrow$ collected images $\{ I_k^{t} \}_{k=1}^{K}$

    \STATE \textit{// Perception Agent}
    \STATE $\mathcal{I}^{t} \leftarrow$ annotate $\mathcal{I}^{t}$ with unique instance IDs from $\mathcal{C}^{t}$
    \STATE Update $a_i$ in $\mathcal{T}$ with perceived attributes based on $\mathcal{I}^{t}$
    \IF{attribute consistent \AND spatial conformity \AND observation sufficiency}
        \STATE status $\leftarrow$ \textit{Completed}
    \ELSE
        \STATE $t \leftarrow t+1$
        \STATE $\mathcal{F}^{t} \leftarrow$ feedback on erroneous candidates and guidance for subsequent planning
    \ENDIF
\ENDWHILE

\STATE $y \leftarrow$ final answer to $Q$ based on $\mathcal{C}^t$ and $\mathcal{I}^{t}$
\RETURN $y$

\end{algorithmic}
\end{algorithm}

\begin{table}[H]
    \centering
    \caption{An example prompt for guiding an MLLM to identify key objects from the textual query description, and extract their corresponding categories and attributes.}
    \small

    \renewcommand{\arraystretch}{1.2}
    \setlength{\tabcolsep}{10pt}
    \begin{tcolorbox}[
        enhanced, 
        colback=gray!5!white, 
        colframe=gray!70!black,
        fonttitle=\bfseries,
        fontupper=\linespread{1.1}\selectfont,   
        title={Prompt Template for Key Objects Identification},
        before skip=0pt, after skip=0pt,
    ]
    \textbf{You are an assistant designed to identify key objects from a 3D scene based on the user's query.}\\
    A \textbf{\textcolor{sm}{key object}} is any object that is explicitly mentioned or implicitly described in the query. For each key object, you must extract:
    \begin{itemize}
        \item \texttt{"Category"}: the object \textbf{{\textcolor{sm}{category}}} as a short noun phrase (e.g., \texttt{"towel"}, \texttt{"stove"}, \texttt{"entrance door"}, \texttt{"wall"}).
        
        \item \texttt{"Attribute"}: a textual description of the object's \textbf{\textcolor{sm}{properties}} that are directly stated in the query (e.g., color, material, or other modifiers). If no attribute is specified for an object, set \texttt{"Attribute": ""}.
    \end{itemize}
    
    \textbf{Here are some examples:}
    \begin{itemize}
      \item \texttt{Query: "The white towel hanging off the stove, next to the sink."}\\
          \texttt{Output: [\{"Category": "towel", "Attribute": "white"\}, \{"Category": "stove", "Attribute": ""\}, \{"Category": "sink", "Attribute": ""\}]}
      

      
      \item \texttt{Query: "Red and wooden chair near the window."}\\
          \texttt{Output: [\{"Category": "chair", "Attribute": "red and wooden"\}, \{"Category": "window", "Attribute": ""\}]}
    \end{itemize}
    \ldots \\\\
    Now, based on the query below, extract each \textbf{\textcolor{sm}{key object with its category and attributes}}.\\
    \textbf{Response in the format:}\\
    \texttt{[\{"Category": "<object category>", "Attribute": "<object attribute>"\}, ...]}\\
    \textbf{User's query:} ``\texttt{Cabinet touching the wall farthest from the entrance door}''.
    \end{tcolorbox}
    \label{tab:supp_key_object}
\end{table}

\begin{table}[H]
    \centering
    \caption{An example prompt for guiding Planning Agent to plan viewpoints for observing key objects, based on the textual query description and requirement, to capture both object attributes and spatial relationships.}
    \small

    \renewcommand{\arraystretch}{1.2}
    \setlength{\tabcolsep}{10pt}
    \begin{tcolorbox}[
        enhanced, 
        colback=gray!5!white, 
        colframe=gray!70!black,
        title style={draw=none},
        fonttitle=\bfseries,
        fontupper=\linespread{1.1}\selectfont,
        title={Prompt Template for Viewpoint Planning},
        before skip=0pt, after skip=0pt,
    ]
    \textbf{You are a spatial reasoning assistant tasked with planning viewpoints based on the user's query.}\\
    Imagine that you are observing the entire scene from a top-down perspective. Your task is to \textbf{\textcolor{sm}{plan viewpoints}} that help locate the target object from the user's query \textbf{\textcolor{sm}{description and requirement}}.  Follow the steps below:\\
    \textbf{Step 1:} Observe the room layout from the BEV and understand its structure.\\
    \textbf{Step 2:} Based on the query, infer which object(s) it refers to ,by \textbf{\textcolor{sm}{considering proximity}}, \textbf{\textcolor{sm}{object type}}, and \textbf{\textcolor{sm}{spatial relations}}.\\
    \textbf{Step 3:} Provide \textbf{\textcolor{sm}{8 diverse camera viewpoints}} that offer grounding evidence for the target object(s): some viewpoints should focus on observing \textbf{\textcolor{sm}{object attributes}}, while others should capture \textbf{\textcolor{sm}{spatial relationships}} among objects.\\\\
    Now, based on the following query description, requirements, and key objects below, provide viewpoints.\\
    \textbf{Response in the format:} \\
    ``\texttt{\{"Cameras": [\{"Position": [x1, y1], "Look\_at": [x\_obj, y\_obj], "Visible\_ids": [id1, id2, ...], "Reasoning": "Explain briefly why this position and target are chosen to help ground the query."\}, \ldots]\}}''.\\
    \textbf{User's query:} ``\texttt{Cabinet touching the wall farthest from the entrance brown door}''.\\
    \textbf{User's requirement:} ``\texttt{Need co-visible viewpoint to capture spatial relationships among ID1, ID3 ...  and nearby viewpoint to verify the color of ID3...}''.\\
    \textbf{Key object IDs and their coordinates information are as follows:} \\
    \texttt{Object ID}: $1$, \texttt{Type}: cabinet, \texttt{Color}: brown, \texttt{Shape}: rectangular, \texttt{Center Coordinates}: $(11, 6)$; \\ 
    \texttt{Object ID}: $3$, \texttt{Type}: door, \texttt{Material}: wooden, \texttt{Shape}: rectangular, \texttt{Center Coordinates}: $(8, 2)$;
    \texttt{Object ID}: $8$, \texttt{Type}: cabinet, \texttt{Shape}: rectangular, \texttt{Center Coordinates}: $(12, 14)$;\\ 
    \ldots
    \end{tcolorbox}
    \label{tab:supp_planning_agent}
\end{table}

\begin{table}[H]
    \centering
    \caption{An example prompt for guiding Perception Agent to document the 3D environment and generate answer, based on the textual query description, object spatial and attribute information.}
    \small
    
    \renewcommand{\arraystretch}{1.2}
    \setlength{\tabcolsep}{10pt}
    \begin{tcolorbox}[
        enhanced, 
        colback=gray!5!white, 
        colframe=gray!70!black,
        title style={draw=none},
        fonttitle=\bfseries,
        fontupper=\linespread{1.1}\selectfont,
        title={Prompt Template for Scene Perception},
        before skip=0pt, after skip=0pt,
    ]
    \textbf{You are an assistant designed to identify the target object based on user's query description.}\\
    You are given \textbf{\textcolor{sm}{real images, rendered images}} and textual 3D information of scenes. Prioritize \textbf{\textcolor{sm}{extracting object attributes}} from real images and \textbf{\textcolor{sm}{inferring inter-object relationships}} from rendered images. Each object is labeled with a unique visible ID.\\
    Your task is to \textbf{\textcolor{sm}{document the object attributes}} and \textbf{\textcolor{sm}{identify the target object}}. For each visual attribute (Color, Material, Shape), you must also estimate a confidence score in the range $[0,1]$. 
    \begin{itemize}
        \item If such a target object exists, directly output the \textbf{\textcolor{sm}{predicted object ID}}.
        \item If no suitable target is found, output the extracted visual attributes (with confidences) of the candidate objects and additionally provide \textbf{\textcolor{sm}{feedback}}, including: which object IDs are incorrect, which parts of the query are not satisfied, what additional information would help.\\
    \end{itemize}

    \textbf{Response in the format:}\\
    If a matching target \textbf{exists}, respond in the format:\\
    ``\texttt{\{"ID": <id1>\}}''.\\
    If a matching target \textbf{does not exist}, respond in the format:\\
    ``\texttt{\{"Attributes": [\{"ID": <id1>, "Color": "<color or unknown>", "Color\_conf": <float in [0,1]>, "Material": "<material or unknown>", "Material\_conf": <float in [0,1]>, "Shape": "<shape or unknown>" "Shape\_conf": <float in [0,1]>\}, \ldots], "Feedback": [\{"Error\_ID": [<id\_error1>, <id\_error2>, \ldots>], "Guidance": "Explain what additional information would be needed."\} ]\}}''.\\
    \textbf{User's query:} ``\texttt{Cabinet touching the wall farthest from the door}''.\\
    \textbf{Key objects with their 3D information are as follows:} \\
    \texttt{Object ID}: $1$, \texttt{Type}: cabinet, \texttt{Color}: brown, \texttt{Color\_conf}: 0.9, \texttt{Shape}: rectangular, \texttt{Shape\_conf}: 0.95, \texttt{Dimensions}: Width $0.21$, Length $3.72$, Height $0.84$, \texttt{Center Coordinates}: X $1.42$, Y $0.76$, Z $0.41$; \\ 
    \texttt{Object ID}: $3$, \texttt{Type}: door, \texttt{Material}: wooden, \texttt{Material\_conf}: 0.85, \texttt{Shape}: rectangular, \texttt{Shape\_conf}: 0.92, \texttt{Dimensions}: Width $1.56$, Length $1.13$, Height $2.30$, \texttt{Center Coordinates}: X $0.61$, Y $3.23$, Z $1.18$; \\
    \ldots
    \end{tcolorbox}
    \label{tab:supp_perception_agent}
\end{table}

\subsection{Additional Results with Multiple Agents}

\begin{table}[t]
    \centering
    \caption{Performance comparison with multiple perception agents on the Nr3D~\cite{achlioptas2020referit3d} validation subset proposed by VLM-Grounder~\cite{xu2024vlm}.
    }
    \setlength{\tabcolsep}{8pt}
    \resizebox{\linewidth}{!}{
    \begin{tabular}{lcccccc}
        \toprule 
        Method & Agent  & Easy & Hard & Dep. & Indep. & Overall \\ 
        \midrule\midrule
        VLM-Grounder~\cite{xu2024vlm}  & GPT-4V~\cite{achiam2023gpt}
        & 55.2 & 39.5 & 45.8 & 49.4 & 48.0 \\
        SeeGround~\cite{li2025seeground}& Qwen2-VL-72B~\cite{qwen2vl} & 51.5 & 37.7 & 44.8 & 45.5 & 45.2 \\
        SeqVLM~\cite{lin2025seqvlm}& Doubao-1.5-vision-pro~\cite{bytedance2025volcengine} & 58.1 & 47.4 & 51.0 & 54.5 & 53.2 \\
        SPAZER~\cite{jin2025spazer}& GPT-4o~\cite{hurst2024gpt} & 68.0 & 58.8 & 59.9 & 66.2 & 63.8 \\
        \rowcolor{resultcolor}
        Ours (1 Perception Agent)& GPT-4o~\cite{hurst2024gpt} & 69.9 & 60.5 & 61.5 & 68.2 & 65.6 \\
        \rowcolor{resultcolor}
        Ours (3 Perception Agent)& Qwen2.5-VL-32B~\cite{bai2025qwen2} & \textbf{60.3} & \textbf{52.6} & \textbf{54.2} & \textbf{58.4} & \textbf{56.8}  \\
        \rowcolor{resultcolor}
        Ours (3 Perception Agent)& GPT-4o~\cite{hurst2024gpt} & \textbf{70.6} & \textbf{62.3} & \textbf{62.5} & \textbf{69.5} & \textbf{66.8}  \\
        \bottomrule
    \end{tabular}}

    \label{tab:app-subset}
\end{table}

\begin{figure}[t]
\centering
\includegraphics[width=0.9\linewidth]{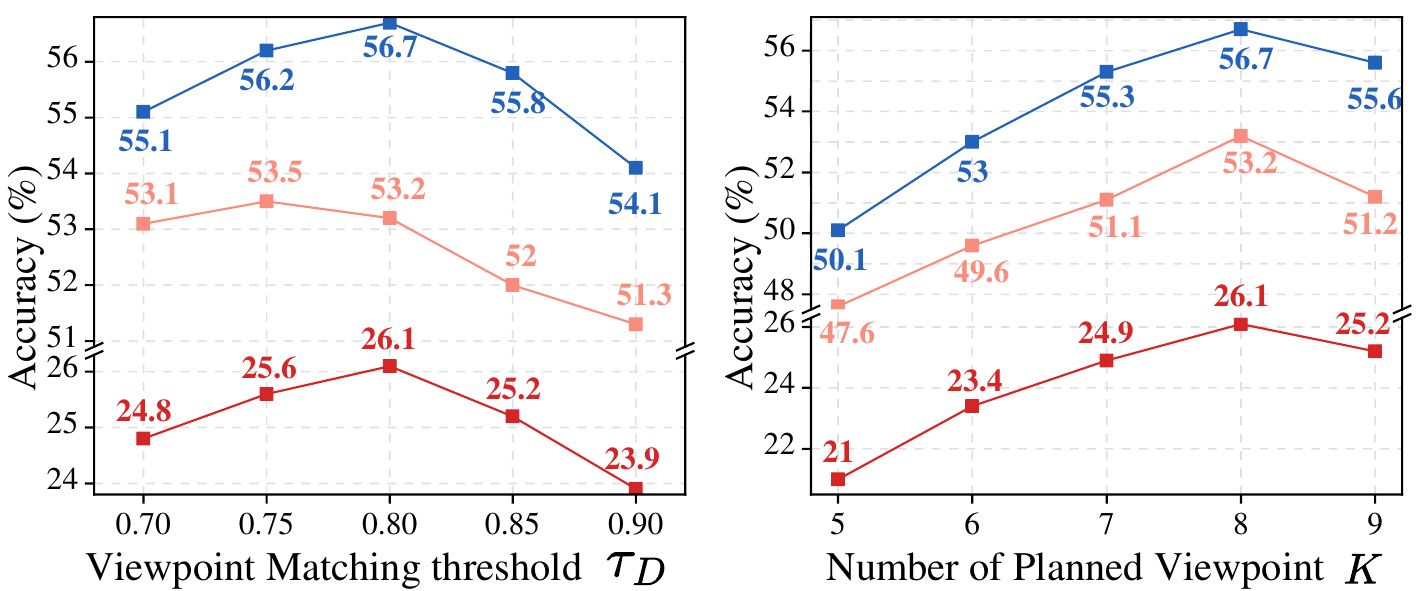} 
\caption{Left: Influence of the viewpoint matching threshold $\tau_D$. Right: Influence of the number of planned viewpoints $K$.
}
\label{fig:viewpoint_number}
\end{figure}

As shown in \cref{tab:app-subset}, increasing the number of Perception Agents to 3 yields further performance improvements, with Qwen2.5-VL-72B~\cite{bai2025qwen2} as Planning Agent and GPT-4o (gpt-4o-2024-08-06~\cite{hurst2024gpt}) as all Perception Agents. Our method outperforms previous state-of-the-art zero-shot method SPAZER~\cite{jin2025spazer} by 3.0\% Acc@0.25 on Nr3D~\cite{achlioptas2020referit3d}, demonstrating the effectiveness and scalability of our multi-agent framework. Moreover, we find that our framework allows smaller MLLMs to achieve strong performance, with Qwen2.5-VL-32B even outperforming most zero-shot methods based on larger MLLMs.

\subsection{Sensitivity Analysis of Hyperparameters}

\noindent\textbf{Effect of Viewpoint Matching Threshold $\tau_D$.}
As demonstrated in the left of \cref{fig:viewpoint_number}, we investigate the effect of viewpoint matching threshold $\tau_D$ across different datasets. As $\tau_D$ increases, performance steadily improves, peaking at $0.75$/$0.8$ as more real images are provided to capture fine-grained semantics. However, as $\tau_D$ further increases, performance degrades as fewer rendered views lead to missing critical perspectives for spatial localization. 

\noindent\textbf{Effect of Number of Planned Viewpoints $K$.}
As shown in the right of \cref{fig:viewpoint_number}, we analyze how varying the number of planned viewpoints $K$ across three benchmarks. As $K$ increases, the accuracy steadily improves, reaching its maximum at $K=8$. This indicates that providing an appropriate number of viewpoints supplies Perception Agent with more comprehensive scene information, allowing a more reliable observation of the target object. However, when $K$ increases from 8 to 9, the accuracy begins to drop. This is because an excessive number of planned viewpoints increases the burden on Planning Agent, resulting in lower-quality viewpoint planning. More importantly, the increased overlap among views introduces redundant information that hampers the inference of Perception Agent.





\subsection{Full Comparison}
We present a comprehensive comparison with previous approaches on ScanQA~\cite{azuma2022scanqa} validation set. As shown in \cref{tab:supp_scanQA_full}, our method substantially outperforms existing zero-shot approaches across all metrics. Compared with  SpatialPrompting~\cite{taguchi2025spatialprompting}, it achieves improvements of 3.4 CIDEr and 1.4\% EM.

\begin{table}[t]
    \centering
    \caption{Full comparison of 3D question answering on ScanQA~\cite{azuma2022scanqa} validation set. ``EM'' denotes top-1 exact match.}
    \resizebox{\linewidth}{!}{
    \begin{tabular}{lcccccccc}
    \toprule
    \multirow{2}{*}{Method}
      & \multicolumn{4}{c}{BLEU-n Metrics}
      & \multicolumn{3}{c}{Language Generation Metrics}
      & \\  
    \cmidrule(r){2-5}  \cmidrule(r){6-8}
    & BLEU-1 & BLEU-2 & BLEU-3 & BLEU-4 & CIDEr & METEOR & ROUGE & EM \\ 
    \midrule\midrule
        \multicolumn{9}{l}{\textit{Supervision: Fully Supervised}}\\
        ScanQA~\cite{azuma2022scanqa}      & 30.7 & 21.2 & 15.8 & 10.8 & 60.2 & 12.6 & 31.1 & 20.9 \\
        3D-LLM~\cite{hong20233d}           & 39.3 & 25.2 & 18.4 & 12.0 & 69.4 & 14.5 & 35.7 & 20.5 \\
        LEO~\cite{huang2024embodied}       & - & - & - & 13.2 & 101.4 & 20.0 & 49.2 & 24.5 \\
        SIG3D~\cite{man2024situational}    & 39.5 & - & - & 12.4 & 68.8 & 13.4 & 35.9 & -    \\
        Chat-Scene~\cite{huang2024chat}    & 43. & 29.1 & 20.6 & 14.3 & 87.7 & 18.0 & 41.6 & 21.6 \\
        Scene-LLM~\cite{fu2025scene}       & 43.6 & 26.8 & 19.1 & 12.0 & 80.0 & 16.6 & 40.0 & 27.2 \\
        Video-3D LLM~\cite{zheng2025video} & 47.1 & 31.7 & 22.8 & 16.1 & 102.1 & 19.8 & 49.0 & 30.1 \\
        3DRS~\cite{huang2025mllms}         & 48.4 & 32.7 & 23.8 & 17.2 & 104.8 & 20.5 & 49.8 & 30.3 \\
        Ross3D~\cite{wang2025ross3d}       & 49.2 & 33.7 & 24.9 & 17.9 & 107.0 & 20.9 & 50.7 & 30.8 \\
    \midrule
        \multicolumn{9}{l}{\textit{Supervision: Zero-shot}}\\
        Agent3D-Zero~\cite{zhang2024agent3d}       & 28.6 & - & - & 4.4 & 71.8 & 16.0 & 37.0 & 17.5  \\
        SpatialPrompting~\cite{taguchi2025spatialprompting} & 38.5 & 25.1 & 16.6 & 10.9 & 87.7 & 16.9 & 43.4 & 27.3  \\
        \rowcolor{resultcolor}
        {Ours} & \textbf{40.0} & \textbf{25.7} & \textbf{17.6} & \textbf{12.3} & \textbf{91.1} & \textbf{17.4} & \textbf{44.5} & \textbf{28.7}  \\
        \bottomrule
    \end{tabular}}
    \label{tab:supp_scanQA_full}
\end{table}

\subsection{Additional Qualitative Results}

\begin{figure*}[t]
\centering
\includegraphics[width=0.9\textwidth]{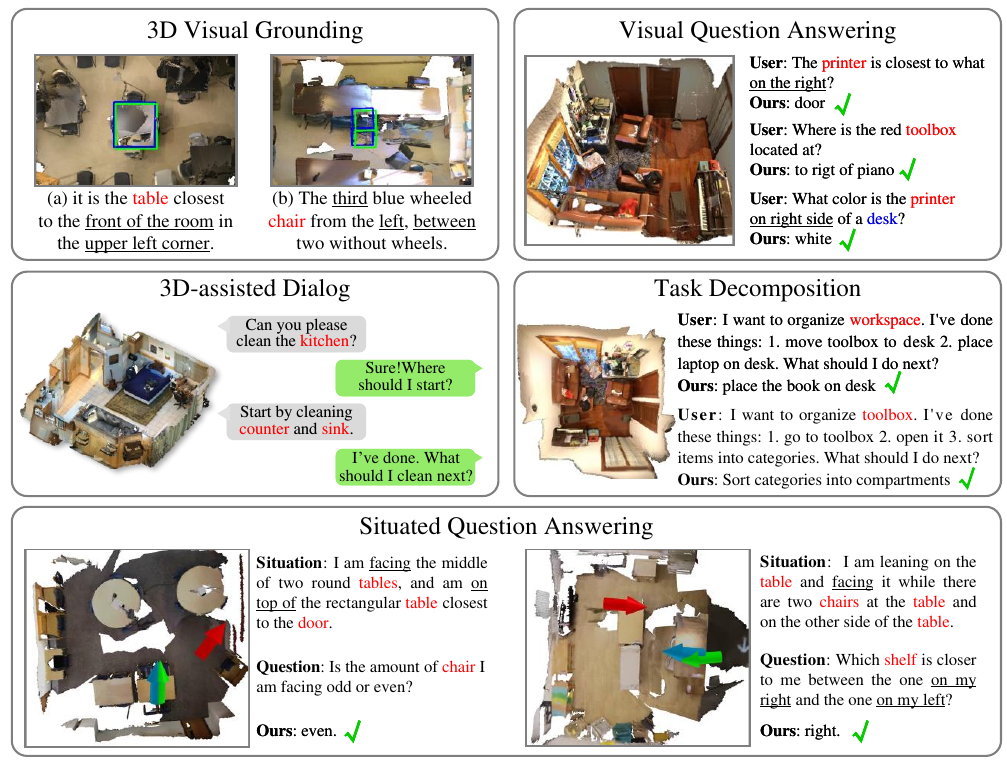}
\caption{Additional qualitative results across various 3D understanding tasks.
}
\label{fig:Qualitative}
\end{figure*}

\cref{fig:Qualitative} provides additional qualitative results, demonstrating the effectiveness of our method across diverse 3D understanding tasks. In complex 3D scenes containing multiple identical objects (\textit{e.g.}, “tables”, “chairs”) and involving intricate textual descriptions, our method can accurately interpret inter-object relationships (\textit{e.g.}, “between”, “on the right”), fine-grained semantics (\textit{e.g.}, “without wheels”, “rectangular”), spatial layouts (\textit{e.g.}, “front of the room”), and egocentric situations (\textit{e.g.}, “facing”).


\section{Limitations and Broader Impact}
\label{secs_supp:limitations_broader_impact}

\begin{wrapfigure}{r}{0.3\linewidth} 
  \centering
  \includegraphics[width=\linewidth]{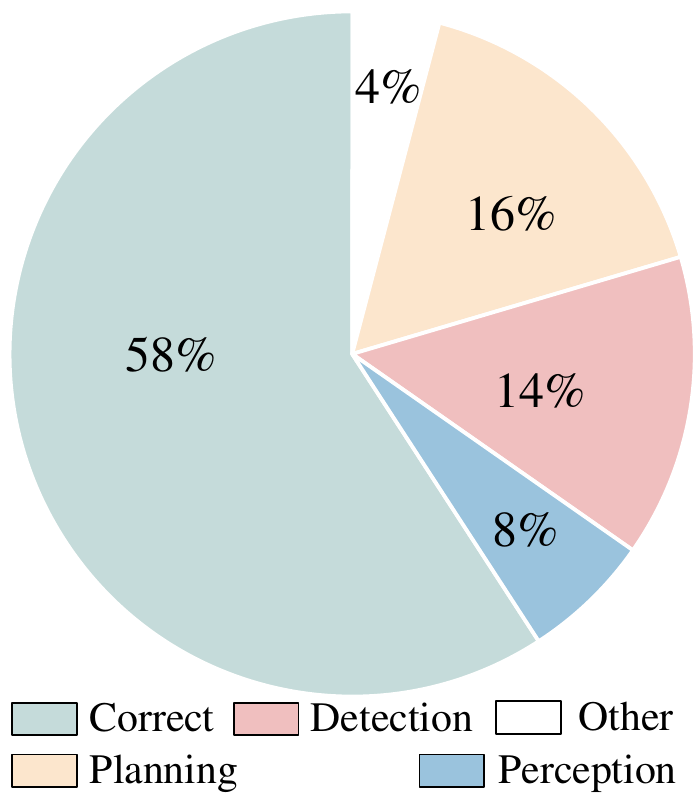}
  \caption{Error type distribution on ScanRefer~\cite{chen2020scanrefer}.}
  \label{fig:error_distribution}
\end{wrapfigure}

\subsection{Error Type analysis}
In \cref{fig:error_distribution}, we present the distribution of different error types to provide insights into potential directions for further improvement. The main error types include: 
1) Detection: The 3D detector fails to detect the target object or predicts the incorrect category, causing subsequent processing to rely on an inaccurate set of candidate objects. 2) Planning: Errors caused by the model fails to understand 3D layout and plan informative viewpoints relevant to the query. 3) Perception: Error where the model fails to correctly interpret spatial relationships or misses key semantic cues even with correctly planned observations. 4) Other: Mainly due to referring ambiguities in the query text, where multiple objects could satisfy the query, as well as a small number of errors caused by incorrect interaction stopping.

We observe that planning and perception errors constitute a major portion, highlighting that the primary challenge of 3D scene understanding lies in informative view selection and spatial-semantic perception. To address this, we plan to further explore more effective viewpoint planning strategy and improve the model’s capability for joint spatial-semantic perception across multi-view observations. Moreover, a considerable portion of errors is caused by the detector, indicating that reducing the reliance on the 3D detector is another important direction for future work.

\subsection{Analysis of Failure Cases}

\begin{figure*}[t]
\centering
\includegraphics[width=0.9\textwidth]{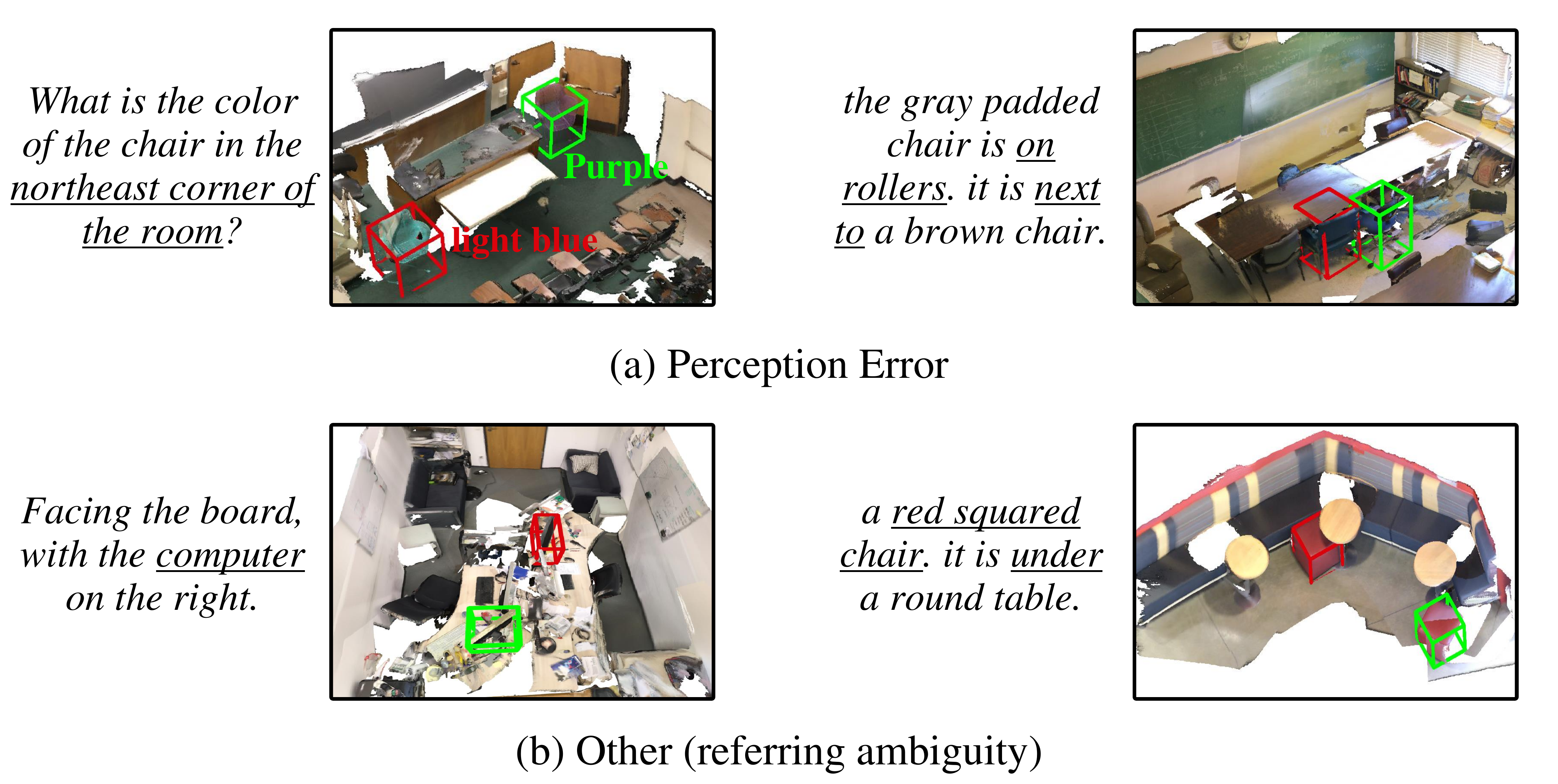}
\caption{Representative failure cases. \textcolor{visualred}{Red} and \textcolor{visualgreen}{green} represent the ground-truth and prediction, respectively. (a) Perception error involves relative relation (\textit{e.g.}, ``next to''), absolute relation (\textit{e.g.}, ``northeast corner''), and detailed object attributes (\textit{e.g.}, ``on rollers''). (b) Other errors are primarily caused by the referring ambiguity in the query.}
\label{fig:failure_case}
\end{figure*}

Since we follow prior works~\cite{jin2025spazer, taguchi2025spatialprompting, li2025seeground} to adopt same detection-based paradigm, detection-related errors are currently unavoidable. In the future, we plan to explore how to integrate a high-fidelity online scene reconstruction model to refine detection errors or directly produce localization results. In addition, planning-related errors suggest that future work should further improve the model’s ability to select informative and query-relevant viewpoints. The remaining two error types are further analyzed through the representative cases shown in \cref{fig:failure_case}.

For scene perception, as shown in \cref{fig:failure_case} (a), the main failures include: 1) Complex spatial relationships, including the absolute orientation of the target object and its relations with surrounding objects. In the future, we plan to explore visual prompt designs with orientation information. 2) Detailed semantic attributes, especially subtle properties that are difficult to capture reliably, such as ``on rollers''. In future research, we will incorporate finer-grained part-level attributes and affordance attributes into the cognitive map.

Moreover, the dataset contains some samples with referential ambiguity. As shown in \cref{fig:failure_case} (b), both the prediction result and the ground-truth label can reasonably match the query. Therefore, the construction of higher-quality dataset through stricter data filtering and annotation protocols is a key issue to be addressed in future work.

\subsection{Broader Impact}
Under a unified planning, our system can arrange multiple agents to jointly explore a scene and construct a shared scene representation, and thus has the potential to exert long-term impact on robotics, human–computer interaction, and augmented reality. By coordinating viewpoint allocation and fusing the perceptual of different agents, the system can achieve faster scene understanding, more stable decision making, and fewer errors in applications such as autonomous navigation, assistive manipulation, and automated inspection, where safety and reliability are crucial. Building on these capabilities, the framework may further promote progress in collaborative perception, multi-agent coordination, and human–AI collaboration, enabling multi-agent and humans to jointly construct and maintain a high-level understanding of complex 3D spaces and to operate effectively in more open-ended, long-horizon tasks.

\end{document}